\documentclass{article} 

\usepackage{iclr2027_conference,times}

\usepackage[utf8]{inputenc} 
\usepackage[T1]{fontenc}    
\usepackage{hyperref}       
\usepackage{url}            
\usepackage{booktabs}       
\usepackage{amsfonts}       
\usepackage{nicefrac}       
\usepackage{microtype}      
\usepackage{xcolor}         
\usepackage{graphicx}
\usepackage{longtable} 
\usepackage{array}     
\usepackage{wrapfig}   

\makeatletter
\def\fnum@table{\textbf{\tablename~\thetable.}}
\def\fnum@figure{\textbf{\figurename~\thefigure.}}
\long\def\@makecaption#1#2{%
  \vskip\abovecaptionskip
  \sbox\@tempboxa{#1 #2}%
  \ifdim \wd\@tempboxa >\hsize
    #1 #2\par
  \else
    \global \@minipagefalse
    \hb@xt@\hsize{\hfil\box\@tempboxa\hfil}%
  \fi
  \vskip\belowcaptionskip}
\long\def\LT@makecaption#1#2#3{%
  \LT@mcol\LT@cols c{\hbox to\z@{\hss\parbox[t]\LTcapwidth{%
    \reset@font
    \sbox\@tempboxa{#1{#2 }#3}%
    \ifdim\wd\@tempboxa>\hsize
      #1{#2 }#3%
    \else
      \hbox to\hsize{\hfil\box\@tempboxa\hfil}%
    \fi
    \endgraf\vskip\baselineskip}%
  \hss}}}
\makeatother

\usepackage{amsmath}
\newcommand{\fulldisplayskips}{\abovedisplayshortskip=\abovedisplayskip\belowdisplayshortskip=\belowdisplayskip}
\usepackage{textcomp}
\usepackage{calc} 

\makeatletter
\@ifundefined{c@none}{}{}
\makeatother

\usepackage{newunicodechar}
\newunicodechar{≫}{\ensuremath{\gg}}
\newunicodechar{≪}{\ensuremath{\ll}}
\newunicodechar{→}{\ensuremath{\rightarrow}}
\newunicodechar{←}{\ensuremath{\leftarrow}}
\newunicodechar{⇒}{\ensuremath{\Rightarrow}}
\newunicodechar{≈}{\ensuremath{\approx}}
\newunicodechar{−}{\ensuremath{-}}
\newunicodechar{÷}{\ensuremath{\div}}
\newunicodechar{…}{\ensuremath{\ldots}}

\makeatletter
\let\iclr@origunderscore\_
\renewcommand{\_}{\iclr@origunderscore\allowbreak}
\makeatother

\usepackage{fvextra}
\fvset{breaklines=true,breakanywhere=true,breaksymbolleft={},breakindent=0pt}
\DefineVerbatimEnvironment{verbatim}{Verbatim}{}

\makeatletter
\newsavebox\pandoc@box
\newcommand*\pandocbounded[1]{
  \sbox\pandoc@box{#1}%
  \Gscale@div\@tempa{\textheight}{\dimexpr\ht\pandoc@box+\dp\pandoc@box\relax}%
  \Gscale@div\@tempb{\linewidth}{\wd\pandoc@box}%
  \ifdim\@tempb\p@<\@tempa\p@\let\@tempa\@tempb\fi
  \ifdim\@tempa\p@<\p@\scalebox{\@tempa}{\usebox\pandoc@box}%
  \else\usebox{\pandoc@box}%
  \fi%
}
\makeatother
\makeatletter
\def\maxwidth{\ifdim\Gin@nat@width>\linewidth\linewidth\else\Gin@nat@width\fi}
\def\maxheight{\ifdim\Gin@nat@height>\textheight\textheight\else\Gin@nat@height\fi}
\makeatother
\setkeys{Gin}{width=\maxwidth,height=\maxheight,keepaspectratio}
\makeatletter
\def\fps@figure{htbp}
\makeatother

\providecommand{\tightlist}{%
  \setlength{\itemsep}{0pt}\setlength{\parskip}{0pt}}
\title{Large-scale factor analysis shows machine intelligence is only partially
interpretable}

\author{%
  Faiz Ghifari Haznitrama\thanks{Equal contribution.}\\
  School of Computing, KAIST\\
  \texttt{haznitrama@kaist.ac.kr}
 \And
  Afrizal Hasbi Azizy\footnotemark[1]\\
  Independent Researcher\\
  \texttt{afrizalhasbi.azi@gmail.com}
 \And
  Faeyza Rishad Ardi\\
  Independent Researcher\\
  \texttt{faeyza.rishad@gmail.com}
}

\iclrfinalcopy
\begin{document}

\maketitle
\lhead{Preprint}

\begin{abstract}
    A common assumption in language model development is that cognitive
    abilities are organized around a general, domain-free intelligence factor,
    like fluid intelligence in humans. This assumption is rarely tested
    directly, and prior attempts have done so only at a much smaller scale. We
    take a latent variable approach to intelligence in language models, similar
    to how psychometricians study psychological constructs. Performance in every
    specific problem set is influenced by a domain-specific and a
    domain-agnostic latent factor. Using factor analysis as a
    dimension-reduction technique, we analyzed 13,230 published evaluation
    scores covering 1,618 language models across 456 different text-only
    benchmarks. Due to the sparse nature of the dataset, we triangulate our
    analysis across different data densifiers and imputation methods. Across the
    imputations that pass our quality filter, the general factor's share of
    variance in the summed score (\(\omega_h\)) ranges from 0.04 to 0.76, and
    its share of the common variance (ECV) from 0.11 to 0.52. A robust pattern
    across different modes of bias is that 1. Same-subject benchmarks is not
    guaranteed to place closely in vector space, with weak median label cohesion
    2. Benchmarks loading highest on \(G\) tends to be miscellaneous rather than
    thematically coherent, and 3. Standard ``intelligence'' benchmarks do not
    dominate the share of ability variance the general factor explains, and
    there is a lack of evidence that \(G\) is well-proxied by them. We
    hypothesize that targeting a single conceptual ability has little support,
    since the first-order abilities it would have to reach are often partially
    idiosyncratic and not identifiable in practice. This would mean that, from
    public benchmark scores alone, general intelligence is hard to define,
    identify, and target as a single construct in language model development.
    Future experiments should run training regimes that target both
    theoretically abstract general ability and tasks with semantic similarity
    and examine how abilities transfer across test sets of diverse formats.
  \end{abstract}

\section{Introduction}\label{introduction}

The field of artificial intelligence is full of theories about what intelligence
``is'' \citep{chollet2019,morris2023,waterhouse2023}. This is partly because
neural networks tend to overfit their training data, such that their performance
across tasks lacks the generalizability of human cognition \citep{zhang2021}. In
pursuit of this generalizability, models are often implicitly assumed to
organize cognitive abilities according to a causal hierarchy, much like humans,
where a general-purpose, ``content-free'' capability underlies and transfers
across many specific downstream tasks \citep{bommasani2021}. A popular label for
this hierarchy, borrowed from psychometric research, is ``fluid intelligence'',
defined as the ability to perform well on any task regardless of content
\citep{cattell1963}. By definition this construct holds a causal relation, where
an increase of fluid intelligence leads to an increase in the mastery of every
other cognitive task.

The concept of general fluid intelligence is borrowed from human intelligence
psychometrics, which discovers the higher-order structure of cognitive abilities
by applying dimension reduction techniques to a large set of cognitive measures.
A growing body of work in machine learning follows this step by analyzing LLM
benchmark scores to ascertain the nature of ``general intelligence'' in
machines. However, a common weakness across these studies is the relatively
small number of benchmarks or variables, ranging from 6 to 23, most of which
measure ``smartness'' narrowly (mathematics, academic knowledge, common-sense
reasoning) and which risk conflating the extracted capability factor with model
scale \citep{kearns2026}. These studies also tend to fit a human-informed
structure, and do not look for bottom-up patterns specific to LLMs.

Within the context of prior work, our study presents factor analysis of 13,230
evaluation scores from 1,618 models and 456 benchmarks, the widest benchmark
coverage to date, to the best of our knowledge. Our pool of benchmarks, unlike
previous work, covers a highly diverse set of tasks, including those quite
outside of the mainstream. A useful analogy is that reaction time in humans is
correlated with intelligence \citep{kranzler1989}, showing that no matter how
seemingly unimportant a task may be, it may provide information entirely
unintuitive to our subjective understanding. Across this analysis, we find that
factor patterns are only partially interpretable and often incoherent, and that
the evidence for a ``general intelligence'' factor proxied by commonly-targeted
``reasoning'' benchmarks is limited, with the highest loadings more often going
to miscellaneous tasks with no common theme.

\section{Background and Related Work}\label{background-and-related-work}

\subsection{The Psychometric Paradigm}\label{the-psychometric-paradigm}

A common theoretical ground in psychometric measurement is that observable human
behaviors are \textbf{causally} influenced by latent variables internal to an
individual \citep{borsboom2003,borsboom2004}, and \citet{federiakin2025} applies
the same distinction to the design of LLM benchmarks. Individuals differ in some
latent factor, hence individuals differ in some observable outcomes. If there
are no causal latent factors, then the covariance among observed behaviors would
have no source at all, which makes little sense.

Personality and intelligence research are prime examples. \citet{allport1936}
extracted all or most of the words from the English dictionary that can describe
someone's personality, and had a large sample self-report how well each word
describes themselves, hence measuring as much of the set of all possible
personalities \citep{john1988}. \citet{spearman1904} did much the same for
intelligence, collecting the scores of students across school subjects and
finding that their variance overwhelmingly loads on a single wide-breadth latent
variable called the \(G\) factor \citep{jensen2002}, which remains generally
accepted \citep{johnson2004,johnson2008}. Both follow the same paradigm.
Exhaustively measure observable behaviors, subject them to dimension reduction
techniques, and draw theories from the resulting latent factor. Subsequent
research decomposes the hierarchy further, into facets for personality
\citep{lee2018,deyoung2007} and into specific cognitive abilities for
intelligence \citep{schneider2018}.

\subsection{Causal Assumptions in Machine
Intelligence}\label{causal-assumptions-in-machine-intelligence}

Various LLM benchmarks are known to be intercorrelated, though the origin of
this covariance is rarely stated explicitly. We examine the implicit assumption
that benchmark correlations originate from latent variables, and add that
existing paradigms of machine intelligence are implicitly causal. The assumed
precedence of a content-free intelligence \citep{chollet2019} already implies
such a relation, and model development relies on it. In aiming to achieve
``general intelligence'', developers tend to train in a targeted manner.
Reasoning-oriented post-training, for instance, is motivated by the expectation
that improvements on ``pure'' logical tasks will transfer to tool-calling,
long-horizon agentic tasks, and coding \citep{deepseekai2025}. Such an
expectation is coherent only if the targeted ability stands in a causal relation
to the rest, which we state as follows.

\textbf{Definition 1.} {\fulldisplayskips{}Let \(F\) be a set of latent factors
and \(T\) the set of performance scores over all possible tasks:
\[F = \{\, f_i \mid i \in \mathcal{I} \,\}, \qquad G \longrightarrow F \longrightarrow T, \quad F \not\longrightarrow G.\]
That is, \(G\) is a higher-order factor that causally affects the set of latent
factors \(F\), through which it in turn influences task performance. No causal
path runs from \(F\) back to \(G\).}

There is a growing assumption in the field that a subset of \(T\), mostly
assumed to be reasoning, mathematics, and coding, measures \(G\) better than any
other abilities. When a model is fine-tuned to perform better on such tasks, the
implicitly expected transfer happens because training improves \(G\) and,
through it, the factors \(F\):

\textbf{Definition 2.} {\fulldisplayskips{}Let \(F\) be a set of latent factors
and \(T\) the set of performance scores over all possible tasks. Task
performance \(t_i\) is given by a linear predictor:
\[t_i = \lambda_gG + \lambda_1 f_1 + \dots + \lambda_k f_k + \epsilon_i, \qquad i \in \mathcal{I},\]
where \(\lambda_g, \lambda_1, \dots, \lambda_k\) are the task's factor loadings
(standardized regression coefficients) and \(\epsilon_i\) is the task's unique
variance.}

\subsection{Interpreting Structural
Patterns}\label{interpreting-structural-patterns}

Understanding the patterns of benchmark correlation is theoretically relevant to
cognitive science, though this relevance does not concern any one specific
factor loading pattern. Under a small number of benchmarks the discovered
patterns depend on which benchmarks were sampled \citep{major2011}, and under a
large number the missingness pattern becomes too large to trust any single
solution. The correct approach is therefore to examine recurring and large-scale
patterns.

One such pattern is the explanatory power\footnote{We use the phrase
  ``explanatory power'' over ``existence'', as \(G\) is a construct of the
  factor model we fit.} of a general intelligence factor. When a general factor
dominates the explained variance of an EFA result, variance in an LLM's
capability is predominantly caused by a latent \(G\), supporting the idea that a
flexible, ``raw'' intelligence is a substantial component of performance and
that targeting it is a fruitful research program. A weak \(G\) factor instead
means that LLM abilities are strongly specific, so improvements require
brute-force training in as much task diversity as possible, with no
silver-bullet construct that parsimoniously describes ``general intelligence''.
A second pattern is which benchmarks load highest. Under the indifference of the
indicator \citep{spearman1904}, content does not determine loading, so the test
is whether a factor recovered from a small battery survives a wider one.

We read these patterns without assuming what a general factor should look like,
since factor analysis is a bottom-up, theory-free approach. Authors tend to
impose theories of human intelligence on artificial neural networks, as we know
of no better model of intelligence than our own. However, it is problematic
because these networks present an entirely different form of cognitive process
than that of biological minds. Human intelligence research itself began with
factor analysis \citep{spearman1904} and continues to apply bottom-up dimension
reduction even though its theories have been well established for decades.
Therefore, we do not assume any general factor to have a tidy definition, and so
we do not discriminate between benchmarks and try to sample as much of the set
of all possible tasks as we can. We argue that it is a reasonable position to
expect that LLMs work in ways entirely unintuitive to the human mind, and having
no priors whatsoever is the correct way to start our inquiry. Purpose-built
diversity suites such as BIG-bench \citep{srivastava2022} are denser, but they
sample one team's construction of task diversity. We are interested in the
benchmarks the field uses.

\subsection{Related Work}\label{related-work}

Prior work applying factor analysis to benchmark scores is smaller in scale, and
mostly confirmatory. \citet{ilicagignac2024} applied confirmatory factor
analysis to hundreds of models and found that their performance fits a
human-informed causal structure, which repeats the mistake of fitting an
existing theory to model intelligence. A large proportion of their benchmarks
are also variants of MMLU \citep{hendrycks2021}, posing the risk of
common-method bias polluting the model fit. \citet{federiakin2025} likewise fits
a single-factor CFA to the HuggingFace Open LLM Leaderboard. A confirmatory fit
can only test the structure its authors specify in advance. \citet{hardy2026}
fit CFA and generalizability theory to over 4,000 models on the same
leaderboard, and find local dependence among its items, which is the
common-method bias we raise above.

\citet{krakauer2026} instead applies PCA and finds a first principal component
declining from 90\% of variance to 64\% by 2024, interpreted as a ``rotation''
in the \(G\) factor as models outsource reasoning to external tools. PCA does
not partition out systematic from error variance and so cannot be trusted to
isolate a genuine \(G\) factor from noise. Only two studies run EFA.
\citet{burnell2023} found three major factors, but did not do higher-order
factor analysis on the resulting loadings, so they cannot say to what extent
those factors are influenced by a presumable \(G\) factor.
\citet{haznitrama2026} report a unified general factor before showing that a
neuropsychologically-grounded battery reveals cognitive gaps that it otherwise
masks. Narrower still is \citet{holm2024}, where a single factor explains 95\%
of variance across just 8 scenarios in one language. Item response theory,
recently applied to LLM benchmarks \citep{polo2024,zhou2025}, and
full-information maximum likelihood both handle incomplete data natively. IRT
requires per-item responses, which published scores do not report, and FIML
requires every benchmark pair to share observed models, which many pairs at our
sparsity do not.

\section{Methodology}\label{methodology}

In this study we investigate the low-dimensional structure of model benchmark
scores, which comprises distinct but correlated latent factors that each
dominantly affects different clusters of benchmarks, and a \(G\) factor that
accounts for the variances of all benchmarks. Our raw \(1{,}618 \times 456\)
matrix is super-sparse (\textasciitilde1.8\% density) and missing not at random
(MNAR) \citep{rubin1976}, since popular models and popular benchmarks are
observed far more often. We therefore apply several densification and imputation
methods, each with its own biases and assumptions, and triangulate their results
to find a common characteristic.

\subsection{Data Collection}\label{data-collection}

\textbf{Scope.} We limit this study to the evaluation of generative language
models on a set of text-only benchmarks. We define generative language models as
those that accept arbitrary prompts and produce text completions. Encoder-only
classifiers, narrow task-specific systems (dedicated MT/ASR/TTS models), and
undocumented community uploads are excluded. Benchmarks are included if they
have at least one in-scope result row. For each row, we follow the schema from
EveryEvalEver \citep{everyevalever2026} to unify the evaluation results. Full
inclusion and exclusion criteria are given in
\hyperref[inclusion-and-exclusion-criteria]{Appendix~\ref*{inclusion-and-exclusion-criteria}}.

\textbf{Sources.} \hyperref[tab:sources]{Table~\ref*{tab:sources}} gives the
composition of the text-only corpus by source family.
\hyperref[data-source-and-normalization]{Appendix~\ref*{data-source-and-normalization}}
lists every named source and the extraction route used for each.

\begin{center}\small\setlength{\tabcolsep}{4pt}
\begin{minipage}[t]{0.47\textwidth}
\refstepcounter{table}\label{tab:sources}\textbf{Table~\thetable.} Composition of the text-only corpus by source family.\par\vspace{4pt}
\centering
\begin{tabular}{@{}lrr@{}}
\toprule
Source family & Rows & Benchmarks \\
\midrule
Stanford HELM & 4,942 & 138 \\
HF Open LLM Leaderboard & 4,529 & 12 \\
Papers With Code & 1,378 & 151 \\
Other online leaderboards & 953 & 65 \\
Kaggle AI Benchmarks & 844 & 26 \\
Primary papers & 584 & 94 \\
\bottomrule
\end{tabular}
\end{minipage}\hfill
\begin{minipage}[t]{0.49\textwidth}\setlength{\tabcolsep}{3pt}
\refstepcounter{table}\label{tab:matrices}\textbf{Table~\thetable.} Aggregated model $\times$ benchmark matrices, text-only corpus. ``Retained'' is the fraction of observed cells surviving the densifier peel.\par\vspace{4pt}
\centering
\begin{tabular}{@{}llrrr@{}}
\toprule
Densifier & Strategy & Shape & Density & Retained \\
\midrule
raw & Std. & 1,266 $\times$ 404 & 2.2\% & \\
raw & Aggr. & 334 $\times$ 380 & 3.5\% & \\
C & Std. & 670 $\times$ 78 & 13.8\% & 65\% \\
C & Aggr. & 201 $\times$ 102 & 13.6\% & 62\% \\
S & Std. & 667 $\times$ 124 & 10.0\% & 75\% \\
S & Aggr. & 123 $\times$ 293 & 10.1\% & 81\% \\
R & Std. & 174 $\times$ 298 & 11.8\% & 55\% \\
R & Aggr. & 97 $\times$ 310 & 11.7\% & 78\% \\
\bottomrule
\end{tabular}
\end{minipage}
\end{center}

\textbf{Protocol.} Given the large number of fields from the EveryEvalEver
schema, we follow a strict source-verification protocol, meaning each field is
populated only if the verified source explicitly documented it. As an exception,
some fields can be inferred from the record itself using deductive rules with
small risk of error (see
\hyperref[inferred-fields]{Appendix~\ref*{inferred-fields}}). In particular, we
put some focus on obtaining the release date field for both models and
benchmarks, and only accept a release date if it is explicitly documented in the
source (\hyperref[release-dates]{Appendix~\ref*{release-dates}}).

We also handle redundancy at two levels. Redundant duplicate rows are removed
(\hyperref[duplicate-detection-and-integrity-checks]{Appendix~\ref*{duplicate-detection-and-integrity-checks}}),
and near-perfectly correlated benchmark identifiers (version, dialect, or
language splits of one benchmark) are collapsed to one representative, which
removes 47 identifiers and 2,216 rows
(\hyperref[score-redundancy-pruning]{Appendix~\ref*{score-redundancy-pruning}}).

\subsection{Data Processing}\label{data-processing}

\textbf{Deduplication.} Rows surviving duplicate removal for the same (model,
benchmark) pair are averaged into a single score. To keep near-duplicate results
from the same models under different conditions (e.g.~reasoning effort) from
breaking the IID assumption, and to further densify the data, we average rows
under two collapse strategies. The \textbf{standard} strategy is variant-level.
It keeps different version numbers and parameter counts, while collapsing
reasoning effort, knowledge cutoff, etc. The \textbf{aggressive} strategy is
family-level. Every model is collapsed to its base family token (Claude, Llama,
etc.), so all sizes and generations of a family form one row.

\textbf{Metric selection.} 92 of our collected benchmarks were reported under
several metrics. As different metrics are not comparable, we keep one metric
covering the most distinct models, so the widest comparable population survives.
A per-benchmark override list handles cases where coverage alone chooses badly.
Finally, benchmarks observed for only one model are dropped. We sign-flip
benchmarks whose selected metric is lower-is-better (e.g.~error rates) so that a
higher score is consistently better.

\textbf{Densification.} We further improve the data density by greedily peeling
the matrix towards a common target density. We feature three densifiers,
differing by the axis they sacrifice: \textbf{C (column-primary)} drops the
least-observed benchmarks, then any model left empty, retaining famous
benchmarks with wide model coverage. \textbf{R (row-primary)} drops the
least-observed models, then any benchmark left empty, retaining a broad
benchmark set over a small set of heavily-evaluated models, which leaves fewer
models than benchmarks. \textbf{S (symmetric)} drops whichever marginal has the
lowest fill-rate, privileging neither axis. After peeling, models and benchmarks
that have fewer than 3 observed scores are dropped. Columns with zero variance
among observed values are also dropped, since they carry no correlational
signal. The algorithm is given in
\hyperref[densification-algorithm]{Appendix~\ref*{densification-algorithm}}. To
aid sequential imputation (below), we run the peel towards a ladder of target
densities from 50\% down to 10\%, so that the final, sparsest slice still
includes as many different benchmarks as possible.

\textbf{Imputation.} We applied missing data imputation algorithms: SoftImpute
\citep{mazumder2010}, k-NN, and missForest \citep{stekhoven2012}. Method-level
descriptions and implementation details for all of the above are given in
\hyperref[imputation-methods]{Appendix~\ref*{imputation-methods}}. Each imputer
runs down a density ladder of 50\%, 40\%, \dots, 10\%: it is seeded by the
densification algorithm at the densest slice
(\hyperref[densification-algorithm]{Appendix~\ref*{densification-algorithm}}),
and each sparser slice is imputed on the union of its own observations and the
previous slice's completed matrix, with observed cells taking precedence. Out
method of imputing the denser part of a matrix first and using it to fill the
sparser part has precedent in least-squares \citep{zhang2008} and
denoising-autoencoder \citep{ma2020} imputations.

\textbf{Evaluating the imputations.} At each step of the chain, we mask
\textasciitilde20\% of the observed cells as an evaluation set. The mask is
column-stratified, so each benchmark is masked at least once and keeps at least
two training observations. Held-out cells are drawn only from benchmarks that
the step introduces (absent from all denser slices), so each benchmark is graded
exactly once across the chain, and are scored by
\(R^2 = 1 - \text{MSE}/\text{MSE}_{\text{baseline}}\), where the baseline
predicts each column's training mean
(\hyperref[imputation-evaluation]{Appendix~\ref*{imputation-evaluation}}).
\(R^2\) selects the hyperparameters within each method (rank, \(k\), number of
trees, etc.). We report the weighted \(R^2\) (\(\mathrm{wr2}\)): the
introduced-benchmark-weighted mean of the per-step held-out \(R^2\), i.e.~the
exact benchmark-balanced mean held-out \(R^2\) of the final matrix; imputations
whose \(\mathrm{wR^2}\) falls below 0.4 are considered unreliable and not
included in factor analysis.

\subsection{Factor analysis}\label{factor-analysis}

To perform our dimension reduction, we use exploratory factor analysis (EFA)
with the minimum residual estimator and the oblique promax rotation\footnote{Rotation
  gives an interpretable ``simple structure''. Oblique rotations allow the
  factors to correlate, while default eigendecomposition yields orthogonal
  solutions.}, followed by the Schmid-Leiman bifactor transformation
\citep{schmidleiman1957}. This runs factor analysis hierarchically, yielding one
additional factor that influences the rest of the extracted factors, and is
widely used in psychometric research on the \(G\) factor of intelligence
\citep{johnson2004,johnson2008}. The hierarchical step makes it a more
principled choice than interpreting the highest-eigenvalue solution (e.g.,
\citealp{krakauer2026}) as the \(G\) factor.

From the bifactor solution we report two statistics for the \(G\) factor. The
first, \(\omega_h\), is the share of variance in the summed score over all
benchmarks that is due to \(G\). The second, the explained common variance
(ECV), is the share of the common variance, excluding error, that is due to
\(G\) \citep{reise2010}.

\textbf{Definition 3}. Let \(T\) be a matrix of test scores that can be
decomposed into independent additive components due to a \textbf{general factor}
(g), \textbf{specific factors} (s)\footnote{Usually called ``group factors''. We
  avoid the term to prevent confusion with observation grouping.}, and
\textbf{error} (\(\epsilon\)). Then \(\omega_h\) is the estimand
\[\omega_h = \frac{\sigma^2_{G}}{\sigma_T^2}, \qquad \sigma_T^2 = \sigma^2_{G} + \sigma^2_{\mathrm{s}} + \sigma^2_{\epsilon}.\]

In practice, with \(\lambda_{g,j}\) the loading of benchmark \(j\) on \(G\),
\(h_j^2\) its communality and \(R\) the benchmark correlation matrix, we compute
\[\omega_h = \frac{(\sum_j \lambda_{g,j})^2}{\mathbf{1}^\top R\,\mathbf{1}}, \qquad \mathrm{ECV} = \frac{\sum_j \lambda_{g,j}^2}{\sum_j h_j^2}\]
with \texttt{psych::omega} \citep{revelle2009,revelle2024}. \(\omega_h\)
describes the summed score, not any single benchmark: the share of one
benchmark's variance due to \(G\) is \(\lambda_{g,j}^2\), which can be much
lower.

One additional step we do is parallel analysis \citep{horn1965} to select the
number of factor analysis dimensions. To keep wall-clock time tractable we cap
the number of factors extracted at 20.

\subsection{Label Cohesion Analysis}\label{label-cohesion-analysis}

We measure benchmark similarity as cosine distance between benchmarks'
factor-analytic loading vectors, described further in
\hyperref[benchmark-embedding]{Appendix~\ref*{benchmark-embedding}}. Given this
benchmark--benchmark distance matrix, we ask whether each label's members sit
closer together than chance. For a label, let \textbf{within} be their mean
pairwise distance. We compare this to the \textbf{null\_mean}: benchmarks are
binned into quartiles of \(\log_{10}(\text{model coverage})\), and the null
redraws, 2,000 times, a random same-size set matched to the label's members'
coverage-quartile composition, recomputing the same mean-pairwise-distance
statistic each draw. Coverage stratification matters because coverage is itself
correlated with tightness in this factor space. Labels with fewer than four
members, or comprising the entire benchmark set, are not scored. The reported
effect size is:

\[A = 1 − \frac{\text{within}}{\text{null\_mean}}\]

0 means the label's members are no tighter than a random draw, 1 means
effectively identical members, and negative values mean the members are more
spread out than chance.

\textbf{Significance.} The p-value is computed empirically from the same 2,000
permutation draws:

\[p = (1 + \#\{\text{null draws at least as tight as observed}\}) / (1 + 2000)\]

Within each cell, we control the false discovery rate across the labels of one
axis via Benjamini--Hochberg at q = 0.05. A label's reported significant count
(e.g.~``6/9'') is the number of cells in which it passed this FDR-corrected
threshold, out of the cells where it could be scored.

\subsection{Note on triangulation}\label{note-on-triangulation}

We emphasize that our conclusions arise not from one or few specific end-to-end
analysis pipelines, but by triangulating results from different applied methods
and aggregate statistics. Hence our results in the next section and in the
appendix should be understood holistically, highlighting the persistent pattern
that robustly appears across different modes of bias.

\section{Results}\label{results}

\subsection{\texorpdfstring{Variance explained by
\(G\)}{Variance explained by G}}\label{variance-explained-by-g}

Only 6 of the 18 dataset-imputer combinations yield a weighted \(R^2\) above the
0.4 threshold
(\hyperref[imputation-results]{Appendix~\ref*{imputation-results}}), which are
subsequently subject to factor analysis.
\hyperref[tab:point-summaries]{Table~\ref*{tab:point-summaries}} below shows the
point summaries of the factor analyses. The most important statistic here is the
\(\omega_h\), the share of variance in the summed score over all benchmarks that
is due to the general factor. Across the six solutions, \(\omega_h\) ranges from
0.04 to 0.76 and ECV from 0.11 to 0.52.

The range of \(\omega_h\) is wide. The best-performing imputer, SoftImpute on C
Std., yielded a \(G\) factor that accounts for 36.7\% of the summed-score
variance. Newer models score higher on the \(G\) factor
(\hyperref[release-date-analysis]{Appendix~\ref*{release-date-analysis}}).
Leaving out single benchmarks shows that the change in \(\omega_h\) is only
weakly related to how often a benchmark is observed (mean \(|r| = 0.241\),
\hyperref[omega-sensitivity]{Appendix~\ref*{omega-sensitivity}}).

\begin{longtable}{@{}llrrrrr@{}}
\caption{Point summaries of factor analysis results. AVE = average variance explained per factor, $k$ = number of factors extracted, $\phi_\text{avg}$ = average inter-factor correlation.}\label{tab:point-summaries}\\
\toprule
Dataset & Imputer & $\text{AVE}^{\uparrow}$ & ${\omega_h}^{\uparrow}$ & $\text{ECV}^{\uparrow}$ & $\phi_\text{avg}$ & ${wR^2}^{\uparrow}$ \\
\midrule
\endfirsthead
\caption[]{(continued)}\\
\toprule
Dataset & Imputer & $\text{AVE}^{\uparrow}$ & ${\omega_h}^{\uparrow}$ & $\text{ECV}^{\uparrow}$ & $\phi_\text{avg}$ & ${wR^2}^{\uparrow}$ \\
\midrule
\endhead
\bottomrule
\endlastfoot
S Std. & missForest & 17.6\% & 0.760 & 0.501 & 0.215 & 0.455 \\
C Std. & k-NN & 9.7\% & 0.731 & 0.451 & 0.216 & 0.442 \\
C Std. & missForest & 22.0\% & 0.713 & 0.520 & -0.088 & 0.566 \\
C Std. & SoftImpute & 9.6\% & 0.367 & 0.126 & 0.038 & 0.586 \\
R Aggr. & SoftImpute & 4.9\% & 0.287 & 0.321 & 0.028 & 0.441 \\
S Std. & SoftImpute & 11.6\% & 0.044 & 0.114 & 0.041 & 0.428 \\
\end{longtable}

\subsection{Benchmark clusters}\label{benchmark-clusters}

\begin{wraptable}{R}{0.48\textwidth}
\vspace{-2pt}
\footnotesize\setlength{\tabcolsep}{3pt}
\centering
\caption{Cohesion of subject labels. Sig. = number of cells in which the label is significant after Benjamini--Hochberg correction. Full cohesion results for every label can be seen in \hyperref[label-cohesion-results]{Appendix~\ref*{label-cohesion-results}}.}\label{tab:cohesion-subset}
\begin{tabular}{@{}lrr@{}}
\toprule
Label & Median A & Sig. \\
\midrule
\texttt{reading\_comprehension} & +0.331 & 3/9 \\
\texttt{commonsense} & +0.329 & 2/9 \\
\texttt{code} & +0.278 & 4/9 \\
\texttt{math} & +0.028 & 0/9 \\
\bottomrule
\end{tabular}
\vspace{4pt}
\end{wraptable}

Figure 1 below shows an illustrative UMAP plot of benchmarks using composite
distances aggregated from factor loadings, colored based on their subject matter
(\hyperref[benchmark-embedding]{Appendix~\ref*{benchmark-embedding}}). In this
plot, benchmarks with a common subject only occasionally cluster together. This
is supported by the low cohesion scores in
\hyperref[tab:cohesion-subset]{Table~\ref*{tab:cohesion-subset}}, where
\texttt{reading\_comprehension}, \texttt{commonsense}, and \texttt{code} show
weak cohesion that is significant in at most 4 of 9 cells, while \texttt{math}
shows no structure distinguishable from chance. Narrow labels such as
\texttt{finance} are cohesive, but they have very few members
(\hyperref[label-cohesion-results]{Appendix~\ref*{label-cohesion-results}}).
Across the entire figure, the spaces occupied by each flagged subject matter
span across the entire plot. In other words, evidence from benchmark
correlations suggest that \textbf{capability in one task does not always
generalize to another task of the same subject}. A semantically coherent
generalization is probable but not guaranteed, which makes domain abilities
difficult to isolate from a purely semantic and intuitive standpoint. This
phenomenon, where same-domain benchmarks lack a tendency to cluster together, is
observed in nearly all of our imputations, which we discuss further in
\hyperref[common-subject-distances]{Appendix~\ref*{common-subject-distances}}.

\begin{figure}
\centering
\pandocbounded{\includegraphics[keepaspectratio]{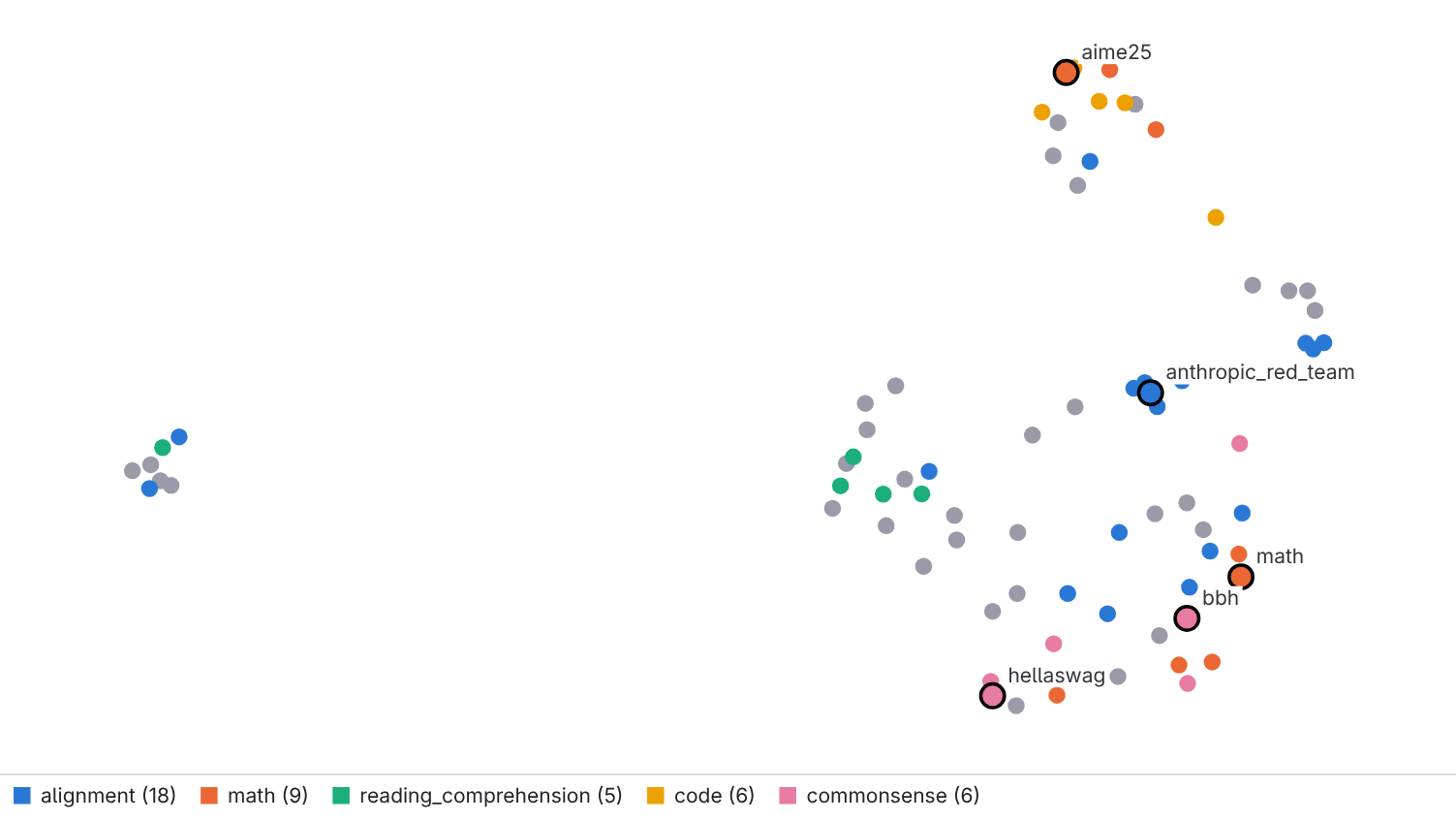}}
\caption{UMAP plot of benchmark distances from C, SoftImpute (\(wR^2 = 0.586\)).
Domain-similar benchmarks are not guaranteed to cluster together. For results
from all our other valid imputations, see
\hyperref[common-subject-distances]{Appendix~\ref*{common-subject-distances}}}
\end{figure}

\subsection{\texorpdfstring{Benchmarks'
\(G\)-centrality}{Benchmarks' G-centrality}}\label{benchmarks-g-centrality}

Another point of interest for the research question is which benchmarks act as a
good proxy of general intelligence, particularly as research is concerned with
performance in certain specific benchmarks to quantify intelligence
advancements. \hyperref[tab:g-rankings]{Table~\ref*{tab:g-rankings}} answers
this question by ranking the benchmarks by the normalized average rank
order\footnote{We use rank order as factor loadings vary in range, and they are
  normalized as different datasets have different numbers of benchmarks.} of
their loadings on the \(G\) factor (\(\rho\)). To account for confounding
effects from benchmark frequency, we report the residuals of the rank order
regressed on frequency (\(\rho_\epsilon\)). Further details and justification
are given in
\hyperref[benchmark-g-rankings]{Appendix~\ref*{benchmark-g-rankings}}.

The top benchmarks are not dominated by common standard benchmarks. The top
proxies include factual question answering (\texttt{simpleqa}, FACTS), medical
exams (\texttt{medqa}), Thai school exams, and bias and safety tests
(\texttt{bbq}). Of the common reasoning benchmarks, only
\texttt{pwc\_arc\_challenge} (rank 8) reaches the top 10. To add to this,
standard intelligence benchmarks like \texttt{mmlu} and \texttt{gpqa\_diamond}
are placed near the middle of the rankings (ranks 46 and 75 of 124; \texttt{gsm}
is 58), with \texttt{swe\_bench} and \texttt{hellaswag} further down (ranks 85
and 116). The benchmarks that load most on \(G\) are therefore not mainly
abstract reasoning tasks. \textbf{In this corpus, reasoning, mathematics, and
coding benchmarks are not the best proxies of the \(G\) factor}, contrary to a
common assumption.

\begingroup\small\setlength{\tabcolsep}{3pt}
\begin{longtable}{@{}rlrrlrrr@{}}
\caption{Benchmarks and their average normalized rank order ($\rho$) of their $G$ factor loadings. Sorted by frequency-residualized rank ($\rho_\epsilon$). $\rho$ ranges from 0 to 1, where 0 = ranked first, 1 = ranked last. $N$ = number of EFA estimations with that benchmark. CI and Best/Worst refer to $\rho$.}\label{tab:g-rankings}\\
\toprule
No & Benchmark & ${\rho_\epsilon}^{\downarrow}$ & $\rho^{\downarrow}$ & 95\% CI & Best & Worst & $N$ \\
\midrule
\endfirsthead
\caption[]{(continued)}\\
\toprule
No & Benchmark & ${\rho_\epsilon}^{\downarrow}$ & $\rho^{\downarrow}$ & 95\% CI & Best & Worst & $N$ \\
\midrule
\endhead
\bottomrule
\endlastfoot
1 & simpleqa & -0.312 & 0.186 & [-0.199, 0.572] & 0.041 & 0.350 & 3 \\
2 & kaggle\_yulongt\_facts\_parametric & -0.307 & 0.189 & [-0.491, 0.870] & 0.006 & 0.504 & 3 \\
3 & kaggle\_andrewmingwang\_asset\_ops\_bench & -0.267 & 0.228 & [-0.424, 0.879] & 0.049 & 0.528 & 3 \\
4 & kaggle\_olamysiak\_eclektic & -0.242 & 0.256 & [0.049, 0.463] & 0.175 & 0.341 & 3 \\
5 & tydiqa & -0.235 & 0.260 & [0.046, 0.474] & 0.195 & 0.358 & 3 \\
6 & facts\_search & -0.232 & 0.262 & [-0.208, 0.732] & 0.104 & 0.472 & 3 \\
7 & medqa & -0.192 & 0.316 & [0.016, 0.615] & 0.065 & 0.819 & 6 \\
8 & pwc\_arc\_challenge & -0.186 & 0.351 & [0.083, 0.619] & 0.104 & 0.767 & 6 \\
9 & thai\_exam\_a\_level & -0.175 & 0.354 & [0.020, 0.688] & 0.123 & 0.959 & 6 \\
10 & bbq & -0.162 & 0.349 & [-0.013, 0.711] & 0.013 & 0.967 & 6 \\
46 & mmlu & -0.070 & 0.367 & [0.027, 0.706] & 0.039 & 0.829 & 6 \\
58 & gsm & -0.040 & 0.449 & [0.022, 0.877] & 0.000 & 0.935 & 6 \\
75 & gpqa\_diamond & 0.019 & 0.548 & [0.353, 0.743] & 0.340 & 0.870 & 6 \\
85 & swe\_bench & 0.043 & 0.535 & [-0.512, 1.581] & 0.084 & 0.919 & 3 \\
116 & hellaswag & 0.162 & 0.599 & [0.329, 0.870] & 0.252 & 0.961 & 6 \\
\end{longtable}
\endgroup

\section{Discussion}\label{discussion}

\subsection{Uninterpretable hierarchical
structure}\label{uninterpretable-hierarchical-structure}

Classically, cognitive sciences have approached intelligence from verbal,
conceptual beginnings \citep{legg2007}. Intelligence is given a definition, then
systems are developed according to the specs provided by the definition. This is
not true in psychometric paradigms, which start by analyzing the covariance of
performance measurements. Thus, adopting the psychometric approach for LLM
benchmarks, our study finds that the strength of the \(G\) factor cannot be
estimated consistently from these data. Adding to this, tasks with similar
contents, like mathematics and coding, do not necessarily cluster together. In
this corpus, benchmark scores do not follow a clearly interpretable structure.

Unlike theories of human psychology \citep{schneider2018} where intelligence is
explicitly assumed to possess higher-order structure, with fluid intelligence as
one higher-order factor that has theoretical causal effects
\citep{vandermaas2014} on the performance of other cognitive abilities,
abilities in LLMs like abstract reasoning, domain memorization, and agentic tool
use, appear in this corpus to be horizontal with respect to each other with no
clear higher-order structure.

\subsection{Related experimental evidence}\label{related-experimental-evidence}

Factor analysis assumes a reflective data generating process, but it is strictly
an analysis of correlations. The next question is whether such causality can be
observed and exploited in practice. One way is to follow psychological science,
where training participants in one cognitive task leads to observable
improvements in a different task (hence, generalization) \citep{simons2016}. For
LLMs, this means measuring a model's performance in coding, fine-tuning it to be
better at mathematics, and then measuring its coding performance again. There
are prior studies of this sort which documents non-obvious ability transfers.
Empirically, we categorize them into 3 sorts in relation to this hypothesis,
described below.

\subsubsection{Low-level primitive ability}\label{low-level-primitive-ability}

Prior experiments have provided some support for this hypothesis, describing
that ability transfers occur through subtle shared representations.
\citet{maniparambil2026when} ran a training experiment on OLMo3-7B where
post-training exclusively on constraint-satisfaction puzzles raised pass@32 on
OlymMATH-Hard from 16.0\% to 36.0\%, suggesting that transferable reasoning
primitives rather than mathematical domain content underlie the gain.
\citet{waheed2025on} found across 3,331 controlled fine-tuning experiments that
syntactic style shapes downstream task gains in non-semantic ways, where
Python-style training favors natural-language reasoning, while Java/Rust-style
training favors mathematics.

\subsubsection{Different-domain transfer}\label{different-domain-transfer}

Other documented transfers appear unpredictable from semantic domain alone.
\citet{liu2025spiral} found that training Qwen3-4B-Base solely on Kuhn Poker
self-play raised mathematical reasoning by 8.6 percentage points and general
reasoning by 8.4 percentage points. \citet{fan2026nexttoken} trained a reward
model on 11M tokens of math-focused web text and found overall RewardBench 2
accuracy rose by 7.6--8.3 percentage points over initialization, with gains of
up to +16.1 on math subsets and consistent improvements on out-of-domain safety
benchmarks. \citet{chung2026insightemb} trained an embedding model exclusively
on mathematical reasoning data and found it transferred to ALFWorld, WebShop,
and ScienceWorld agents, outperforming in-domain retrievers despite zero
environment-specific training.

\subsubsection{Same-domain forgetting}\label{same-domain-forgetting}

Third, training on similar domains can actively hurt performance.
\citet{dymkiewicz2026donors} found that single-source LoRA fine-tuning yields a
mean change of −0.75 percentage points across tasks, with degradations typically
larger than gains. The interference is directional, where adding one reasoning
task to a training mix can lower accuracy on another task from the same corpus
by 9--12 percentage points \citep{siboni2026helps}. Within coding, joint
training on React, Vue, and Angular causes compilation failures because the
model mixes their syntax \citep{ma2026phasewise}, and providing a Python
reference for Cangjie code drops pass rates by 19.5 percentage points as the
model adopts Python idioms the other compiler rejects \citep{cheng2026can}. Even
two agent benchmarks collide: training on household-agent tasks slashes accuracy
on another agent benchmark from 41.5\% to 6.6\% \citep{kong2026agentic}. Shared
labels like ``coding'' or ``agent reasoning'\,' do not prevent negative
transfer.

\subsubsection{Relating observational and experimental
evidence}\label{relating-observational-and-experimental-evidence}

In light of the experimental evidence, it is possible to.
Table\textasciitilde{}\ref*{tab:transfer-mapping} makes the correspondence
explicit.

\begin{longtable}{@{}p{0.32\textwidth}p{0.64\textwidth}@{}}
\caption{Correspondence between experimental transfer findings and our observational results.}\label{tab:transfer-mapping}\\
\toprule
Experimental finding & Our observations \\
\midrule
\endfirsthead
\caption[]{(continued)}\\
\toprule
Experimental finding & Our observations \\
\midrule
\endhead
\bottomrule
\endlastfoot
Transfer through low-level task representations & $G$ explains only a substantive share of variance ($\omega_h$ 0.04--0.76, ECV 0.11--0.52), but is domain-free \\
\midrule
Transfer from unrelated semantic domains & Miscellaneous benchmarks load highest on $G$; \texttt{mmlu} and \texttt{gpqa\_diamond} rank mid-table (46th and 75th of 124) \\
\midrule
Negative transfer within similar domains & Same-subject benchmarks do not cluster (median cohesion $A$ near chance for \texttt{math}) \\
\end{longtable}

With regards to this correspondence, several issues remain open. For one, there
is a lack of evidence that singular low-level task representations can account
for the universe of all (or at least, most) language model abilities. Moreover,
it's not clear how fine-grained this low-level representation is. Available
experimental evidence are limited to a few set of benchmarks with unknown
reporting bias. Are reasoning-like primitives homogenous? If not, are they
hierarchically structured such that there is a single reasoning-like latent
variable that accounts for all of their variants? Our analysis suggests no to
both, given how benchmarks rank on the \(G\) loadings. On a similar vein,
negative transfers of same-domain training should only exist if models do not
group task representations by their semantic interpretability.

Nevertheless, factor analysis by itself is a strictly statistical observational
procedure. Stronger evidence and a more final view of machine intelligence
should be acquired by running empirical training experiments that 1. Tests a
considerably higher number of benchmarks, 2. Analyze the heterogeneity of
low-level representations (e.g., are there distinct reasoning-like primitives?),
3. Maps how tasks what kind of shared representation do tasks share with each
other such that they are correlated. Such an experiment would be challenging to
do, however, as there is no reliable way (as our observations suggest) to
determine how many low-level representation there are, how correlated each of
them are, and to what degree they are fundamental to model abilities.

\section{Conclusion}\label{conclusion}

We asked how the cognitive abilities of language models can be organized around
a general factor. Our answer comes from factor analyses of 13,230 published
evaluation scores covering 1,618 models and 456 benchmarks, triangulated so that
no single treatment of the missing data carries the finding. Across the
imputations that pass our quality filter, a general latent factor accounts for
4\% to 76\% of the variance in the summed score (\(\omega_h\) from 0.04 to 0.76)
and 11\% to 52\% of the common variance (ECV). Domain-similar benchmarks are
only occasionally located close together in vector space, and benchmarks with
high \(G\)-loading do not share a common theme and differ from the benchmarks
the field treats as standard measures of intelligence. We conclude that a
hierarchical structure of machine intelligence, as commonly but only implicitly
assumed in the field, is not clearly identifiable from this corpus, given the
hard-to-interpret factor structure. Correlations arise from unintuitive and
unpredictable common features. Whether a non-hierarchical network of abilities
describes these data better is left for future work.

\bibliography{iclr2027_conference}
\bibliographystyle{iclr2027_conference}

\appendix
\renewcommand{\thetable}{A\arabic{table}}
\setcounter{table}{0}
\renewcommand{\thefigure}{A\arabic{figure}}
\setcounter{figure}{0}

\section{Limitations}\label{limitations}

All scores in the corpus are as published, and we evaluate no model ourselves.
This means we do not control the evaluation conditions behind any score, and we
make no attempt to correct for differences in undocumented evaluation setup
between sources. Where two sources disagree about the same evaluation, we
average the two scores. Release dates enter only the exploratory analysis of
\hyperref[release-date-analysis]{Appendix~\ref*{release-date-analysis}}, and
none of the factor analyses depends on them.

Not every completion method we implement is carried through the full design. The
results reported here cover the imputers in
\hyperref[tab:point-summaries]{Table~\ref*{tab:point-summaries}}. The
correlation-level estimators were dropped before the sequential imputation,
since none reached the quality filter at 10\% density
(\hyperref[imputation-methods]{Appendix~\ref*{imputation-methods}}), and
regularized EM-PCA is excluded entirely because its built-in cross-validation is
intractable at this matrix size.

\section{The Use of Large Language
Models}\label{the-use-of-large-language-models}

We used language models to explore the feasibility of and options for our
methodological designs, but the end decisions are ultimately the authors'. For
writing, AI helped us draft parts of the paper and adjust some of the prose, but
every part of the text was checked, validated and further polished by the
authors. AI also assisted in writing the code of the pipeline, under our full
supervision and review, and we verified its numerical output against the
underlying data. Lastly, we used locally-run language models to extract
evaluation results and metadata from leaderboards, papers, repositories and
model cards, since there was far too much to read by hand. Every extraction was
checked by the authors (the release-date checks are reported in
\hyperref[release-dates]{Appendix~\ref*{release-dates}}), and we take full
responsibility for all of the data, as well as every claim, number and citation
in this paper.

\section{Data source and normalization}\label{data-source-and-normalization}

\subsection{Data sources}\label{data-sources}

The corpus is assembled from published evaluation records, and for every score
we record which of four source tiers it comes from.

\textbf{Tier 1, curated evaluation suites} with a standardized harness and one
evaluator across many models. These are Stanford HELM (the Classic, Lite,
Safety, Reasoning, MedHELM, SEA-HELM, Arabic, ThaiExam, EWoK, TORR and Finance
leaderboards) and the HuggingFace Open LLM Leaderboard (v1 and v2).

\textbf{Tier 2, aggregators and result trackers.} Papers With Code, Kaggle AI
Benchmarks, llm-stats.com, Artificial Analysis, Vellum, LiveBench, Chatbot Arena
/ LMArena and pricepertoken.com.

\textbf{Tier 3, benchmark-specific leaderboards}, about 30 in total
(e.g.~BigCodeBench, CRUXEval, SWE-bench, BFCL, VMLU, SEA-LION, PubMedQA and
AlpacaEval).

\textbf{Tier 4, primary papers} reporting original evaluations (arXiv, ACL
Anthology, OpenReview, ACM Digital Library and journals). These are also our
source for benchmark metadata.

The ``Other online leaderboards'' row of
\hyperref[tab:sources]{Table~\ref*{tab:sources}} combines Chatbot Arena /
LMArena (202 rows), llm-stats.com (114), Vellum (85), Artificial Analysis (77),
LiveBench (55) and the remaining named leaderboards (420). For 295 rows (2.2\%)
the source organization was not recorded, and we assigned these by source name
and URL host.

\subsection{Extraction routes}\label{extraction-routes}

\begin{itemize}
\tightlist
\item
  \textbf{Papers With Code.} The public API is defunct (the domain redirects to
  HuggingFace). Evaluation tables are instead read from a daily-published
  archive of the evaluation tables, in four shards. Each row is one \emph{task}
  with nested datasets, each carrying its own leaderboard. Extraction flattens
  task → dataset → leaderboard row into result rows, generating benchmark
  identifiers under their own namespace and applying the scope filter
  (\hyperref[benchmarks]{Appendix~\ref*{benchmarks}}) to exclude non-LLM tasks.
\item
  \textbf{Kaggle AI Benchmarks.} Community-maintained leaderboard datasets,
  extracted per dataset owner and namespaced by owner as well as by benchmark.
  We retain the owner namespace precisely so that cross-source duplicates remain
  detectable, and several were subsequently detected
  (\hyperref[score-redundancy-pruning]{Appendix~\ref*{score-redundancy-pruning}}).
\item
  \textbf{Stanford HELM.} Extracted per sub-project into staging files, then
  merged. We stage before merging so that a partial or malformed extraction can
  be discarded without touching the canonical tables.
\item
  \textbf{Papers and repositories.} ArXiv PDFs and abstracts converted to HTML
  for methodology extraction. GitHub READMEs and evaluation scripts read
  directly from the raw file host, under both default-branch names, and
  HuggingFace dataset cards through the Hub API.
\end{itemize}

\subsection{Inferred fields}\label{inferred-fields}

Two fields are filled from the record itself when the source leaves them blank.

\begin{enumerate}
\def\labelenumi{\arabic{enumi}.}
\tightlist
\item
  \textbf{Inference platform of closed models.} A closed-weights model cannot
  have been run locally, so we set its inference platform to a vendor API.
\item
  \textbf{Temperature under the evaluation harness.} Generative tasks in
  EleutherAI's evaluation harness default to greedy decoding, so for benchmarks
  documented to use the harness we record a sampling temperature of zero.
\end{enumerate}

\subsection{Release dates}\label{release-dates}

We record the release date of every model and benchmark to the month, together
with the kind of evidence behind it. For models, the direct sources are the
creation date of a HuggingFace repository named after the model (801 models), a
web page that we fetched and checked to name the model and give the date (434),
the model's own paper (25) and a date written in the model name (1). These date
1,261 of the 1,618 models (78\%). Most of the rest come from a cited page that
we could not re-read (210) or from a single language-model lookup (89), and 5
models are undated. Benchmark dates are weaker. Only one benchmark is dated from
a direct source (its arXiv identifier), and most carry either a date from an
earlier pass whose origin was not recorded (223) or one that two lookups agree
on only to the year (136).

Every date that came with a citation was checked by fetching the cited page and
asking whether it names the model and gives the date. Of the 565 such dates, 404
were confirmed and 96 could not be checked, since the page was blocked or
paywalled. The other 65 were re-checked by hand against other sources, and 61 of
them were resolved this way.

Two caveats remain. A repository is often created some time before the model is
made public, so its creation date is a lower bound on the release. Also, dates
from language-model lookups tend to be too early for models released after the
training cutoff of the lookup model. Correcting the weaker tiers moved 119 model
dates, 85 of them to a later month, but the dates still resting on a single
lookup may run early.

\subsection{Inclusion and exclusion
criteria}\label{inclusion-and-exclusion-criteria}

\subsubsection{Models}\label{models}

\textbf{Included.} We include general-purpose generative LLMs, domain- or
task-adapted models (code, medical, legal) that still accept arbitrary prompts,
and multimodal models built by adding an encoder to an LLM backbone, provided
the backbone still handles arbitrary text prompts.

\textbf{Excluded.} We exclude encoder-only or classification-only models, narrow
single-purpose systems that cannot be prompted generally (such as dedicated
translation or speech systems), bare embedding or vision encoders,
non-deployable research systems, evaluation metrics misfiled as models, and
undocumented community uploads whose source cannot be traced.

\textbf{Not a separate model.} A different setup of the same model, such as a
context-length variant, a reasoning mode, an effort level or a prompting scheme,
is recorded on the result row instead.

Borderline cases, such as encoder-decoder models, are reviewed one by one
against the same criterion, namely whether the model can follow an arbitrary
prompt and generate text.

\subsubsection{Benchmarks}\label{benchmarks}

We keep only benchmarks that do not require image, audio, video or speech
understanding, and remove benchmarks that are left with no results once
out-of-scope models are removed. The content of a benchmark is not otherwise
filtered.

\subsubsection{Benchmark translation
duplicates}\label{benchmark-translation-duplicates}

We remove benchmarks that are literal translations of another benchmark in the
corpus, and keep those whose content is sourced separately for each language,
checking each case against the benchmark's paper. The removed ones are the
per-language variants of MGSM and Global MMLU Lite, a translated Arabic MMLU,
the Multilingual MMLU and Global MMLU aggregates, IndicXNLI and HumanEval-XL.
Translated benchmarks with no original in the corpus (the language splits of
XCOPA, XNLI and XQuAD) are merged into one benchmark each. Natively multilingual
benchmarks such as MultiLoKo, ArabicMMLU, FLORES, LINDSEA and the Thai national
exams are kept. We also keep cross-language aggregates (MGSM, Belebele and
MultiLoKo), since none of them duplicates a column we hold (MGSM shares no
models with GSM8K, for instance) and they also measure transfer across
languages.

\subsection{Duplicate detection and integrity
checks}\label{duplicate-detection-and-integrity-checks}

Two rows are duplicates if they agree on model, benchmark, metric, setup,
source, model identifier and language, and redundant copies are removed after
review. Rows that differ in any of these are separate evaluations and are kept,
and all rows of one model-benchmark pair are averaged at aggregation (87
model-benchmark-metric triples have scores from more than one source). After
every write we check that every result row points to an existing model and
benchmark and that no model or benchmark is left without results, and we flag
benchmarks with fewer than five rows for review.

\subsection{Score normalization}\label{score-normalization}

Scores are normalized to a 0 to 100 scale. Papers With Code and Kaggle report
scores in mixed formats. For these two sources, a raw value in \([0,1]\) is
multiplied by 100 and a value above 1 is kept as it is. The other sources report
on one scale per leaderboard and are converted as a whole. A few columns remain
on a 0 to 1 scale, which column standardization absorbs. Results are capped at
100 to absorb floating-point noise. Exempt metrics, kept on their native scale,
are perplexity, bits-per-byte, BLEURT, BERTScore, Elo, and count-type metrics
(``\# eval'').

\subsection{Canonical metric selection}\label{canonical-metric-selection}

When a benchmark is reported under more than one metric, we keep one. Metric
names are first normalized for case and whitespace, and spelling variants of the
same measurement (such as written-out and abbreviated forms of accuracy or bits
per byte) are merged by a hand-curated alias map. We then keep the metric that
covers the most models, unless a per-benchmark override applies, and break ties
by row count and then by name. This affects 92 benchmarks and drops 1,420 rows
(706 model-benchmark cells), and about half of these benchmarks lose no model.

Accuracy and exact match are kept apart, even though they look like the same
measure on multiple-choice tasks. On the sixteen benchmarks that carry both,
exact match comes from HELM and accuracy mostly from the Open LLM Leaderboard
and papers, and no model is scored both ways, so the offset between the two
cannot be estimated. Merging them would put two evaluation regimes into one
column. The coverage rule picks accuracy on four of these benchmarks (MMLU,
TruthfulQA, HellaSwag and PubMedQA) and exact match on the other twelve.

\subsection{Remaining column defects}\label{remaining-column-defects}

Two columns need fixing after metric selection. On GPQA, 447 of 454 rows are the
normalized accuracy of the Open LLM Leaderboard v2 (chance mapped to zero, with
negative values clamped to zero), and the other 7 are raw accuracy from papers
and llm-stats.com. The two ranges do not overlap, so we drop the 7 raw rows.
Since 56 of the remaining rows sit at the clamp, GPQA separates weak models
poorly. On ELEPHANT the metric field holds model configurations instead of
metrics, so we remove the benchmark (9 models).

Three other benchmarks (WildBench, SEA-Exam and MultiPL-E) have sources with
non-overlapping score ranges. We leave them as they are, since trackers of
frontier models evaluate stronger models and a gap alone does not show a scale
conflict.

\section{Score-redundancy pruning}\label{score-redundancy-pruning}

This pass is applied to the text-only copy, after the scope filter of
\hyperref[benchmarks]{Appendix~\ref*{benchmarks}}. Each family was audited by
computing the full pairwise Pearson correlation among its columns over the
models evaluated on both columns. The removal decision was taken per family, on
the evidence, and we verified that the resulting cascades orphan no models.

\begin{longtable}{@{}>{\raggedright\arraybackslash}p{0.19\textwidth}>{\raggedright\arraybackslash}p{0.32\textwidth}>{\raggedright\arraybackslash}p{0.20\textwidth}r@{}}
\caption{Benchmark families audited for score redundancy, with the correlation evidence and decision for each.}\label{tab:a3}\\
\toprule
Family & Correlation evidence & Decision & Rows removed \\
\midrule
\endfirsthead
\caption[]{(continued)}\\
\toprule
Family & Correlation evidence & Decision & Rows removed \\
\midrule
\endhead
\bottomrule
\endlastfoot
LiveCodeBench release windows v1--v6 (Kaggle) & mean pairwise $r = 0.995$, worst pair $0.987$, over 45 shared models & Keep the aggregate, drop 6 per-version identifiers & 270 \\
TwitterAAE dialect splits (African-American and White English) & $r = 0.993$--$0.999$ with each other and the parent, over 32 shared models & Keep the parent, drop both dialect splits & 64 \\
GPQA variants (few/zero-shot $\times$ diamond/main, Kaggle) & mean $r = 0.944$, worst pair $0.915$, over 46--47 shared models, with the better-populated canonical GPQA and GPQA Diamond already present & Drop all 4 Kaggle variants & 185 \\
MultiLoKo per-language splits (31 languages, Kaggle) & mean pairwise $r = 0.82$, near-duplicate for well-resourced pairs (Simplified/Traditional Mandarin $0.989$, Italian/Swedish $0.983$) and noisy for low-resource pairs on small overlap & Keep the paper-sourced MultiLoKo across-language aggregate, drop all 31 per-language identifiers & 1,523 \\
A Kaggle re-import of MMLU & Cross-source re-import (44 rows) of the canonical MMLU column (468 rows) & Drop the re-import & 44 \\
A Kaggle re-import of SciCode & Not a third metric. Per-model value matching shows it splices main-problem scores for 30 of 46 models and sub-problem scores for the other 13, which is a scraping artifact & Drop as a data-integrity fix. The 4 explicit split variants ($r = 0.71$--$0.96$) are \textbf{kept}, as their correlations are not uniform enough to treat as duplicates & 46 \\
Stanford HELM ThaiExam sub-splits & Two clusters, not uniform redundancy: \{ONET, IC, A-Level\} at $r = 0.92$--$0.95$, while \{TGAT, TPAT1\} correlate weakly with that cluster ($r = 0.70$--$0.88$). The TGAT/A-Level gap was verified as systematic, not noise (several multilingual models score 35--45 points higher on TGAT) & Drop ONET and IC, and \textbf{keep} A-Level as the knowledge-cluster representative and keep TGAT and TPAT1, which carry distinct variance & 84 \\
 & & \textbf{Total} & \textbf{2,216} \\
\end{longtable}

Two of the seven audited families were therefore kept partly or fully intact.
This pass leaves 456 benchmarks. The canonical-metric filter
(\hyperref[canonical-metric-selection]{Appendix~\ref*{canonical-metric-selection}}),
the source-scale fix and the single-row anomaly removal then drop a further
1,463 result rows, for a final corpus of \textbf{456 benchmarks, 1,618 models,
13,230 result rows}.

\section{Model-identity collapse}\label{model-identity-collapse}

Both strategies operate on the source-specific model identifier, which each
result row carries exactly as its source spelled it, alongside the canonical
model it resolves to. Keeping both is what makes cross-source duplicate
detection possible after canonicalization. Each identifier's family and
parameter-count metadata is first canonicalized to the first non-null value
observed for it. Multiple evaluations of the same (identifier, benchmark) are
averaged before collapsing, and rows sharing a collapse key are averaged again
per benchmark.

\subsection{Standard (variant-level)}\label{standard-variant-level}

We apply the following steps in order to each identifier.

\begin{enumerate}
\def\labelenumi{\arabic{enumi}.}
\tightlist
\item
  Strip the organization prefix, meaning everything before the first slash.
\item
  Reduce parenthesized content to a parameter count where one is present, and
  drop it otherwise.
\item
  Strip trailing separators and any text after them, reasoning-effort phrases
  such as ``medium effort'' or ``high reasoning'', every date format we
  encountered, and any bare four-digit number, which in this corpus is a release
  year, a checkpoint stamp, or a context length.
\item
  Normalize version numbers written with hyphens so that they read as decimals.
\item
  Where a model family is recorded and is not itself numeric, take it as the
  stem and tokenize the remaining suffix. Otherwise tokenize the whole cleaned
  identifier.
\item
  Then, token by token, drop generic training and serving tokens (instruction
  tuning, chat, base, the preference-optimization family, thinking and reasoning
  markers, greedy decoding, and API markers), language and region tokens, legacy
  engine names, and month names. Preserve named tier tokens, since Opus, Sonnet,
  Haiku, Pro, Mini, Flash, Maverick and Scout each denote a distinct released
  model rather than a serving option. Drop context-length tokens. Recognize a
  parameter count either from a billions suffix or by matching the identifier's
  own recorded size, falling back to a whitelist of common sizes. Merge a
  version number into the family stem when one extends the other.
\item
  Emit the family stem, the preserved tokens, and the parameter count.
\end{enumerate}

The common-size whitelist is guarded against version-number collisions, since a
bare 3 or 4 in a Claude or GPT identifier is a version rather than a parameter
count.

\subsection{Aggressive (family-level)}\label{aggressive-family-level}

Take the first alphabetic token of the recorded model family where it is not
numeric. Otherwise strip the organization prefix, the parentheses and the dates
from the identifier, and take its first alphabetic token. Everything else is
discarded.

\subsection{Post-collapse filtering}\label{post-collapse-filtering}

Benchmarks observed for only one collapse key are dropped, then collapse keys
with no remaining benchmarks are dropped. This yields the two raw matrices of
\hyperref[tab:matrices]{Table~\ref*{tab:matrices}}, 1,266 × 404 at 2.2\% and 334
× 380 at 3.5\%, from 2,183 distinct source-level model identifiers.

\section{Densification algorithm}\label{densification-algorithm}

Target density \(\tau = 0.10\). Let \(\mathbf{M}\) be the boolean observation
mask.

\textbf{C (column-primary peel).} While density \(< \tau\): drop the benchmark
(column) with the fewest observations across currently-kept rows, then drop any
model left with zero observations.

\textbf{R (row-primary peel).} While density \(< \tau\): drop the model with the
fewest observations among currently-kept columns, then drop any benchmark left
with zero observations.

\textbf{S (symmetric peel).} While density \(< \tau\): compute each kept
column's and each kept row's \emph{fill rate} (observations ÷ current
opposite-axis size) and drop whichever single marginal has the lowest rate, then
clear emptied rows and columns on both axes.

\textbf{Minimum-observation floor.} After peeling, iterate to a fixed point,
dropping any kept row or column with fewer observations than the floor
\emph{within the currently kept submatrix}. The loop is required because
dropping a sparse row can starve a column and vice versa.

\textbf{Degenerate-column guard.} Finally drop any column with fewer than 2
observed values or zero variance among its observed values. The downstream
estimators would drop these columns anyway, so this keeps the reported matrix
shape equal to the shape that is actually factored.

The tables analyzed in this paper were generated with the floor set to 3
observations.

\section{Imputation methods}\label{imputation-methods}

Every method implemented in R shares one interface: it takes the sparse matrix
and returns the completed matrix, the swept parameter grid, and the held-out
RMSE and \(wR^2\) at each parameter value. None of the methods does any
factoring itself, so the factoring stage is identical across methods.
\hyperref[tab:imputation-cell-methods]{Table~\ref*{tab:imputation-cell-methods}}
lists the full-dataset missing-data estimators.

\begin{longtable}{@{}>{\raggedright\arraybackslash}p{0.15\textwidth}>{\raggedright\arraybackslash}p{0.30\textwidth}>{\raggedright\arraybackslash}p{0.12\textwidth}>{\raggedright\arraybackslash}p{0.26\textwidth}@{}}
\caption{Estimators of the missing dataset entries.}\label{tab:imputation-cell-methods}\\
\toprule
Method & Description & R Package & Configuration \\
\midrule
\endfirsthead
\caption[]{(continued)}\\
\toprule
Method & Description & R Package & Configuration \\
\midrule
\endhead
\bottomrule
\endlastfoot
SoftImpute \citep{mazumder2010} & Nuclear-norm-penalized low-rank completion by iterative soft-thresholded SVD, assuming a low-rank signal plus noise. Primary cell-level method. & softImpute & sweeps rank: 1\ldots10 (capped at $\min(n,p)-1$), and at each rank a 30-point geometric $\lambda$ grid from $\lambda_0$ down to $\lambda_0/100$, ALS with warm starts \\*
k-NN & Each missing cell filled from the $k$ most similar models, an assumption-light baseline with no low-rank, linearity, or normality assumption. & VIM & sweeps $k$: 1\ldots10 (capped below $n$), Gower distance over benchmarks, weighted-mean aggregation \\*
missForest \citep{stekhoven2012} & Iterative random-forest imputation, nonparametric, able to capture nonlinear dependence the low-rank methods cannot represent. & missForest & sweeps number of trees: \{50, 100, 200, 400\}, at most 10 iterations \\*
\end{longtable}

We also tried methods that complete the correlation matrix instead of the data
matrix, including a correlation-level variant of SoftImpute. In a single-step
imputation at 10\% density, none of them reached the quality filter (best
held-out \(R^2\) of 0.39), so we did not carry them into the sequential
imputation and do not factor them.

\section{Imputation evaluation}\label{imputation-evaluation}

\subsection{Sequential imputation}\label{sequential-imputation}

Imputation descends a ladder of target densities from 50\% down to 10\%. The
densest slice is imputed directly from its observations. Each sparser step
trains on the union of its slice's observed cells and the previous step's
completed matrix, with observed cells taking precedence. Because the
densification is deterministic until it reaches the target, higher-density
(smaller subset) slices are guaranteed to be fully nested by a lower-density
(bigger subset) slice.

\subsection{Per-step protocol}\label{per-step-protocol}

At each step, we mask \textasciitilde20\% of the observed cells as an evaluation
set. The mask is column-stratified, so each benchmark is masked at least once
and keeps at least two training observations. Columns are standardized using
training-cell moments only.

Two restrictions keep the graded cells out of training. First, a per-step mask
marks which cells are real observations. Cells carried in from the previous
step's completions are used for training but are never held out or graded.
Second, held-out cells come only from the benchmarks the step introduces, which
are absent from all denser slices. Their cells were never observed, trained on
or completed at an earlier step, so each benchmark is graded exactly once across
the chain, and no step trains on a cell that a later step grades. The
hyperparameter sweep within each step uses the same held-out cells.

Within each benchmark \(j\) introduced at the step, sample
\(n_{\text{hold}} = \max\big(1,\ \min(\lfloor 0.2\, n_{\text{obs}(j)} \rfloor,\; n_{\text{obs}(j)} - 2)\big)\)
observed cells for \(n_{\text{obs}(j)} > 2\), so at least 2 training
observations remain in every column. Every introduced column with more than 2
observations contributes at least one held-out cell.

Let \(\text{MSE}_j\) and \(\text{base}_j\) be the mean squared error and mean
baseline within column \(j\). Then

\[\text{RMSE} = \frac{1}{p}\sum_j \sqrt{\text{MSE}_j}, \qquad R^2 = 1 - \frac{\frac{1}{p}\sum_j \text{MSE}_j}{\frac{1}{p}\sum_j \text{base}_j}\]

\(R^2\) selects the hyperparameters within each method at each step (rank,
number of trees, etc.).

\subsection{Weighted R²}\label{weighted-ruxb2}

The reported evaluation metric is the weighted \(R^2\). Let \(R^2_d\) be the
held-out \(R^2\) of ladder step \(d\) and \(c_d\) the number of benchmarks it
introduces; then

\[\mathrm{wr2} = \frac{\sum_d c_d R^2_d}{\sum_d c_d}\]

Because each benchmark is introduced at exactly one step and graded only there,
\(\mathrm{wr2}\) is the exact benchmark-balanced mean held-out \(R^2\) of the
final matrix, decomposed by introduction step. Factor analysis proceeds only on
imputations whose \(\mathrm{wr2} \geq 0.4\).

\section{Imputation simulation}\label{imputation-simulation}

Held-out evaluation measures how well the masked cells are recovered, but the
analysis reports factor-analytic quantities. To check the full pipeline
(imputation, exploratory factor analysis and the Schmid--Leiman transformation),
we simulate data with a known correlation matrix and factor structure,
calibrated to resemble the real data. We do not expect the pipeline to recover
the planted structure. Instead, we compare the results computed using incomplete
subsets with results using the complete simulated data.

\subsection{Synthetic data generation}\label{synthetic-data-generation}

Two real dataset tables, C (670 \(\times\) 78) and S (667 \(\times\) 124),
supply the observation margins: expected per-row and per-column observation
counts are calibrated to the real tables' counts, so the synthetic data
reproduce the sparsity profile of the corpus under study.

In this simulation, we simulate data using first-order (or, ``suites'') and
second-order factors. For first-order cluster factors, each benchmark loading
strongly on exactly one with deterministic decay so within-suite correlations
pile up near +1. They are the dominant, highest-eigenvalue factors, unlike the
dense background factors on which every benchmark cross-loads. Note that this
multi-order factors are not intended to be a recovery target, but to make data
simulation easier to do. Overall, benchmark \(j\)'s score for model \(i\) is

\[z_{ij} = \sum_{f=1}^{F} \lambda_{jf}\,\eta_{i,f} \;+\; \lambda^{s}_{j}\,\zeta_{i,\mathrm{suite}(j)} \;+\; \sqrt{\psi_j}\,\epsilon_{ij}, \qquad R = L L^\top + S S^\top,\]

with \(F \in \{5, 10\}\) dense background factors whose loadings are drawn
\(U(0.1, 0.5)\) (``weak'') or \(U(0.5, 0.9)\) (``loud'') and scaled down by
factor rank; and \(N_{\text{suites}}\) suite factors, each benchmark loading its
own suite strongly, with a fixed decay per suite, so that the eigenvalues of the
planted factors strictly decrease. All loadings are positive, so \(R\) has only
positive entries. The planted suite membership is the benchmark's dominant
factor.

The model has \(1 + N_{\text{suites}}\) factors: one emergent dominant direction
plus the suite factors. The weak profile (\(U(0.1, 0.5)\)) spreads each
benchmark's variance thinly across many factors, and the loud profile
(\(U(0.5, 0.9)\)) concentrates it in a few. Crossing the two profiles with
\(F \in \{5, 10\}\) covers a plausible range of dimensionality and cross-loading
strength, and we report results per truth cell so that no single structure
favors one method.

The truth contains no general-factor block and no \(\omega_h\) target: the
dominant common direction is emergent, and the only calibrated global quantity
is the mean off-diagonal correlation, planted at \(0.3\) or \(0.5\) and realized
exactly by two block scales. Truth cells are named accordingly:
\texttt{f5\_weak}, \texttt{f5\_loud}, \texttt{f10\_weak}, \texttt{f10\_loud},
crossed with the mean-\(r\) target (\texttt{r30}, \texttt{r50}), giving 8
planted truths per table.

The planted model is one fixed matrix per table, defined over the union of all
density slices. Since the sequential imputation passes each denser slice's
completions on to the sparser ones, the completions at every density have to be
scored against the same complete matrix.

Observation probability follows

\[\mathrm{logit}\, P(\mathrm{obs}_{ij}) = a_i + b_j + s_{ij},\]

where \((a_i, b_j)\) are fit by damped iterated bisection so expected row and
column observation counts match the real table's margins. The \(s_{ij}\) term is
the missingness mechanism:

\begin{longtable}{@{}>{\raggedright\arraybackslash}p{0.18\textwidth}>{\raggedright\arraybackslash}p{0.18\textwidth}>{\raggedright\arraybackslash}p{0.50\textwidth}@{}}
\caption{Description of the simulated data-generation processes}\\
\toprule
Mechanism & $s_{ij}$ & Interpretation  \\
\midrule
\endfirsthead
\caption[]{(continued)}\\
\toprule
Mechanism & $s_{ij}$ & Interpretation  \\
\midrule
\endhead
\bottomrule
\endlastfoot
mcar & $0$ & Calibrated missing-completely-at-random control \\*
cell & $\gamma \cdot z_{ij}$ & Good cells are more likely observed \\*
popularity & $\kappa \cdot \hat\rho_j \, g_i$ & Observability tied to a benchmark's expected correlation, interacting with model ability \\*
\end{longtable}

When a mechanism is too strong for the real margins to be met, we halve its
strength until calibration converges and record the final scale. We use a single
mask seed, and for a given table the random draws behind the masks are the same
across truth cells and mechanisms, so all comparisons between conditions are
paired. In total the grid is 8 truths \(\times\) 3 mechanisms \(\times\) 2
tables = 48 masked datasets.

As in the real corpus, the masks are nested across density slices: each sparser
slice keeps the observed cells of the denser slice and reveals more cells, in
priority order, until its own margins are met. For every mechanism and truth, we
check that the slices are nested (in both their rows and columns and their
observed cells) and that each planted correlation matrix is symmetric, has a
unit diagonal, is positive semi-definite, has strictly decreasing eigenvalues
over the planted factors, and hits its mean-\(r\) target.

\subsection{Protocol}\label{protocol}

Each simulated dataset holds only the C and S tables (standard collapse), and we
run the same sequential imputation as on the real corpus
(\hyperref[imputation-evaluation]{Appendix~\ref*{imputation-evaluation}}).
Starting from the densest slice and descending to the sparsest, each step trains
on its observed cells together with the previous step's completions, with
observed cells taking precedence, and holds out cells only from the benchmarks
the step introduces.

Two levels of scoring follow:

\begin{itemize}
\tightlist
\item
  \textbf{Cell level.} Each step's completions are scored against the complete
  simulated matrix on the missing cells of the introduced columns, balanced by
  column. This \(R^2\) against the planted truth replaces the held-out \(R^2\),
  and the weighted \(R^2\) averages it over steps, weighted by the number of
  columns each step introduces.
\item
  \textbf{Factor level.} The exact factor-analysis recipe (parallel analysis
  factor count, minimum-residual/promax EFA, Schmid--Leiman transformation) runs
  on each completed matrix. A \textbf{fullset} arm runs the identical recipe on
  the complete, unmasked simulated matrix, which separates estimator bias from
  missingness damage. No quality filter is applied at factoring time; the filter
  is applied at scoring so that its effect can be reported.
\end{itemize}

\subsection{Metrics}\label{metrics}

Let \(\omega_{h}^{\text{full}}\) be the fullset arm's general-factor coefficient
for a given (densifier, truth cell) and \(\omega_{h}^{\text{est}}\) the estimate
from a completed matrix. Since no \(\omega\) target is planted, the fullset arm
is the attainable reference, and the reported bias is
\(\Delta\omega_h = |\omega_{h}^{\text{est}} - \omega_{h}^{\text{full}}|\). Runs
enter the comparison only if their final weighted \(R^2\) is at least \(0.4\).

\subsection{Results}\label{results-1}

\hyperref[tab:synth-factors]{Table~\ref*{tab:synth-factors}} lists the runs that
pass the weighted \(R^2 \geq 0.4\) quality filter on the hierarchical end
result, sorted by weighted \(R^2\).

\begin{longtable}{@{}llllrrrr@{}}
\caption{Factor-level scoring on the synthetic corpus, hierarchical end result: runs with weighted $R^2 \geq 0.4$. $\Delta\omega_h$ is bias against the fullset arm.}\label{tab:synth-factors}\\
\toprule
Method & Table & Mask & Truth cell & $\omega_h^{\text{full}}$ & $\omega_h^{\text{est}}$ & $\Delta\omega_h$ & $wR^2$ \\
\midrule
\endfirsthead
\caption[]{(continued)}\\
\toprule
Method & Table & Mask & Truth cell & $\omega_h^{\text{full}}$ & $\omega_h^{\text{est}}$ & $\Delta\omega_h$ & $wR^2$ \\
\midrule
\endhead
\bottomrule
\endlastfoot
missForest & C & popularity & f5\_weak\_r50 & 0.740 & 0.806 & +0.066 & 0.496 \\*
missForest & S & popularity & f10\_weak\_r50 & 0.707 & 0.848 & +0.142 & 0.456 \\*
missForest & C & popularity & f10\_weak\_r50 & 0.730 & 0.856 & +0.126 & 0.450 \\*
SoftImpute & C & mcar & f5\_weak\_r50 & 0.740 & 0.702 & $-$0.038 & 0.449 \\*
missForest & S & popularity & f5\_loud\_r50 & 0.775 & 0.862 & +0.087 & 0.448 \\*
missForest & S & mcar & f10\_weak\_r50 & 0.707 & 0.886 & +0.179 & 0.424 \\*
missForest & S & popularity & f5\_weak\_r50 & 0.693 & 0.804 & +0.111 & 0.415 \\*
missForest & C & popularity & f10\_loud\_r50 & 0.775 & 0.839 & +0.064 & 0.411 \\*
missForest & C & popularity & f5\_loud\_r50 & 0.741 & 0.835 & +0.094 & 0.409 \\*
SoftImpute & C & cell & f5\_weak\_r50 & 0.740 & 0.718 & $-$0.022 & 0.402 \\*
missForest & C & mcar & f5\_weak\_r50 & 0.740 & 0.836 & +0.096 & 0.401 \\
\end{longtable}

All surviving runs are generated with mean-\(r\) \(0.5\); no run generated with
mean-\(r\) \(0.3\) passes the quality filter. missForest accounts for nine of
the eleven surviving runs, spanning both tables and two of the three mechanisms
(no missForest run passes under the cell mechanism); SoftImpute accounts for the
remaining two, both on the C table. Estimated \(\omega_h\) exceeds the fullset
arm's estimate in all nine missForest runs (\(\Delta\omega_h\) between \(0.064\)
and \(0.179\)) and falls slightly below it in both SoftImpute runs (\(-0.022\)
and \(-0.038\)): under missingness, missForest overestimates the general-factor
share relative to the full-data reference, while SoftImpute slightly
underestimates it. The largest bias occurs for missForest on the
\texttt{f10\_weak} cells (\(0.142\) and \(0.179\) under the popularity and mcar
mechanisms respectively), while SoftImpute shows the smallest
(\(0.022\)--\(0.038\)).

\subsection{All runs}\label{synth-all-runs-section}

The table above shows only the runs that pass the quality filter. To show what
the filter selects, \hyperref[tab:synth-summary]{Table~\ref*{tab:synth-summary}}
summarises all 96 runs (2 methods \(\times\) 2 tables \(\times\) 8 truth cells
\(\times\) 3 mechanisms), and
\hyperref[tab:synth-all-runs]{Table~\ref*{tab:synth-all-runs}} lists each run.
One mask seed is used per condition, so the spread reported here is across
conditions, not across repetitions.

\begin{longtable}{@{}llrrr@{}}
\caption{All 96 simulation runs summarised by method and missingness mechanism, split by whether the run passes the weighted $R^2 \geq 0.4$ quality filter. $|\Delta\omega_h|$ = mean absolute difference from the fullset $\omega_h$. Each cell is 16 runs (8 truth cells $\times$ 2 tables), one seed each.}\label{tab:synth-summary}\\
\toprule
Method & Mask & Pass & $|\Delta\omega_h|$ pass & $|\Delta\omega_h|$ fail \\
\midrule
\endhead
\bottomrule
\endlastfoot
missForest & cell & 0/16 & -- & 0.19 \\
missForest & mcar & 2/16 & 0.14 & 0.13 \\
missForest & popularity & 7/16 & 0.10 & 0.11 \\
SoftImpute & cell & 1/16 & 0.02 & 0.05 \\
SoftImpute & mcar & 1/16 & 0.04 & 0.11 \\
SoftImpute & popularity & 0/16 & -- & 0.23 \\
\end{longtable}

The quality filter does not reduce the missForest \(\omega_h\) bias within a
mechanism: missForest overestimates \(\omega_h\) in every passing run.
SoftImpute passing runs have a small \(\omega_h\) error, but SoftImpute never
passes on the S table or under the popularity mechanism, so the simulation does
not directly validate the SoftImpute solution on the real S table. For these
reasons we do not interpret any single real-data \(\omega_h\) value as the
strength of the general factor.

{\footnotesize\setlength{\tabcolsep}{3.5pt}
\begin{longtable}{@{}llllrrrrc@{}}
\caption{Every simulation run. $\Delta\omega_h$ is signed (estimate minus fullset). Pass = weighted $R^2 \geq 0.4$.}\label{tab:synth-all-runs}\\
\toprule
Method & Table & Mask & Truth cell & $\omega_h^{\text{full}}$ & $\omega_h^{\text{est}}$ & $\Delta\omega_h$ & $wR^2$ & Pass \\
\midrule
\endfirsthead
\caption[]{(continued)}\\
\toprule
Method & Table & Mask & Truth cell & $\omega_h^{\text{full}}$ & $\omega_h^{\text{est}}$ & $\Delta\omega_h$ & $wR^2$ & Pass \\
\midrule
\endhead
\bottomrule
\endlastfoot
missForest & C & cell & f10\_loud\_r30 & 0.615 & 0.828 & +0.214 & 0.087 & \\
missForest & C & cell & f10\_loud\_r50 & 0.775 & 0.942 & +0.168 & 0.329 & \\
missForest & C & cell & f10\_weak\_r30 & 0.644 & 0.808 & +0.164 & 0.240 & \\
missForest & C & cell & f10\_weak\_r50 & 0.730 & 0.888 & +0.158 & 0.305 & \\
missForest & C & cell & f5\_loud\_r30 & 0.617 & 0.789 & +0.173 & 0.158 & \\
missForest & C & cell & f5\_loud\_r50 & 0.741 & 0.893 & +0.152 & 0.358 & \\
missForest & C & cell & f5\_weak\_r30 & 0.603 & 0.774 & +0.171 & 0.161 & \\
missForest & C & cell & f5\_weak\_r50 & 0.740 & 0.856 & +0.116 & 0.367 & \\
missForest & C & mcar & f10\_loud\_r30 & 0.615 & 0.796 & +0.181 & 0.105 & \\
missForest & C & mcar & f10\_loud\_r50 & 0.775 & 0.908 & +0.133 & 0.349 & \\
missForest & C & mcar & f10\_weak\_r30 & 0.644 & 0.731 & +0.087 & 0.191 & \\
missForest & C & mcar & f10\_weak\_r50 & 0.730 & 0.910 & +0.180 & 0.362 & \\
missForest & C & mcar & f5\_loud\_r30 & 0.617 & 0.637 & +0.021 & 0.161 & \\
missForest & C & mcar & f5\_loud\_r50 & 0.741 & 0.890 & +0.149 & 0.335 & \\
missForest & C & mcar & f5\_weak\_r30 & 0.603 & 0.700 & +0.097 & 0.171 & \\
missForest & C & mcar & f5\_weak\_r50 & 0.740 & 0.836 & +0.096 & 0.401 & yes \\
missForest & C & popularity & f10\_loud\_r30 & 0.615 & 0.774 & +0.159 & 0.220 & \\
missForest & C & popularity & f10\_loud\_r50 & 0.775 & 0.839 & +0.064 & 0.411 & yes \\
missForest & C & popularity & f10\_weak\_r30 & 0.644 & 0.723 & +0.080 & 0.248 & \\
missForest & C & popularity & f10\_weak\_r50 & 0.730 & 0.856 & +0.126 & 0.450 & yes \\
missForest & C & popularity & f5\_loud\_r30 & 0.617 & 0.443 & -0.174 & 0.187 & \\
missForest & C & popularity & f5\_loud\_r50 & 0.741 & 0.835 & +0.094 & 0.409 & yes \\
missForest & C & popularity & f5\_weak\_r30 & 0.603 & 0.561 & -0.042 & 0.221 & \\
missForest & C & popularity & f5\_weak\_r50 & 0.740 & 0.806 & +0.066 & 0.496 & yes \\
missForest & S & cell & f10\_loud\_r30 & 0.593 & 0.816 & +0.223 & 0.099 & \\
missForest & S & cell & f10\_loud\_r50 & 0.746 & 0.946 & +0.199 & 0.271 & \\
missForest & S & cell & f10\_weak\_r30 & 0.582 & 0.934 & +0.351 & 0.118 & \\
missForest & S & cell & f10\_weak\_r50 & 0.707 & 0.891 & +0.184 & 0.399 & \\
missForest & S & cell & f5\_loud\_r30 & 0.606 & 0.824 & +0.218 & 0.065 & \\
missForest & S & cell & f5\_loud\_r50 & 0.775 & 0.930 & +0.155 & 0.335 & \\
missForest & S & cell & f5\_weak\_r30 & 0.637 & 0.842 & +0.205 & 0.126 & \\
missForest & S & cell & f5\_weak\_r50 & 0.693 & 0.871 & +0.178 & 0.325 & \\
missForest & S & mcar & f10\_loud\_r30 & 0.593 & 0.769 & +0.176 & 0.133 & \\
missForest & S & mcar & f10\_loud\_r50 & 0.746 & 0.863 & +0.117 & 0.345 & \\
missForest & S & mcar & f10\_weak\_r30 & 0.582 & 0.780 & +0.198 & 0.168 & \\
missForest & S & mcar & f10\_weak\_r50 & 0.707 & 0.886 & +0.179 & 0.424 & yes \\
missForest & S & mcar & f5\_loud\_r30 & 0.606 & 0.701 & +0.095 & 0.147 & \\
missForest & S & mcar & f5\_loud\_r50 & 0.775 & 0.905 & +0.130 & 0.361 & \\
missForest & S & mcar & f5\_weak\_r30 & 0.637 & 0.760 & +0.123 & 0.158 & \\
missForest & S & mcar & f5\_weak\_r50 & 0.693 & 0.846 & +0.154 & 0.374 & \\
missForest & S & popularity & f10\_loud\_r30 & 0.593 & 0.751 & +0.157 & 0.175 & \\
missForest & S & popularity & f10\_loud\_r50 & 0.746 & 0.864 & +0.118 & 0.389 & \\
missForest & S & popularity & f10\_weak\_r30 & 0.582 & 0.735 & +0.152 & 0.229 & \\
missForest & S & popularity & f10\_weak\_r50 & 0.707 & 0.848 & +0.142 & 0.456 & yes \\
missForest & S & popularity & f5\_loud\_r30 & 0.606 & 0.618 & +0.012 & 0.220 & \\
missForest & S & popularity & f5\_loud\_r50 & 0.775 & 0.862 & +0.087 & 0.448 & yes \\
missForest & S & popularity & f5\_weak\_r30 & 0.637 & 0.725 & +0.087 & 0.230 & \\
missForest & S & popularity & f5\_weak\_r50 & 0.693 & 0.804 & +0.111 & 0.415 & yes \\
SoftImpute & C & cell & f10\_loud\_r30 & 0.615 & 0.628 & +0.013 & 0.179 & \\
SoftImpute & C & cell & f10\_loud\_r50 & 0.775 & 0.685 & -0.090 & 0.315 & \\
SoftImpute & C & cell & f10\_weak\_r30 & 0.644 & 0.635 & -0.009 & 0.220 & \\
SoftImpute & C & cell & f10\_weak\_r50 & 0.730 & 0.685 & -0.045 & 0.277 & \\
SoftImpute & C & cell & f5\_loud\_r30 & 0.617 & 0.543 & -0.074 & 0.208 & \\
SoftImpute & C & cell & f5\_loud\_r50 & 0.741 & 0.703 & -0.039 & 0.322 & \\
SoftImpute & C & cell & f5\_weak\_r30 & 0.603 & 0.631 & +0.028 & 0.192 & \\
SoftImpute & C & cell & f5\_weak\_r50 & 0.740 & 0.718 & -0.022 & 0.402 & yes \\
SoftImpute & C & mcar & f10\_loud\_r30 & 0.615 & 0.549 & -0.066 & 0.160 & \\
SoftImpute & C & mcar & f10\_loud\_r50 & 0.775 & 0.635 & -0.140 & 0.371 & \\
SoftImpute & C & mcar & f10\_weak\_r30 & 0.644 & 0.546 & -0.097 & 0.284 & \\
SoftImpute & C & mcar & f10\_weak\_r50 & 0.730 & 0.631 & -0.099 & 0.297 & \\
SoftImpute & C & mcar & f5\_loud\_r30 & 0.617 & 0.348 & -0.269 & 0.183 & \\
SoftImpute & C & mcar & f5\_loud\_r50 & 0.741 & 0.628 & -0.113 & 0.249 & \\
SoftImpute & C & mcar & f5\_weak\_r30 & 0.603 & 0.473 & -0.130 & 0.276 & \\
SoftImpute & C & mcar & f5\_weak\_r50 & 0.740 & 0.702 & -0.038 & 0.449 & yes \\
SoftImpute & C & popularity & f10\_loud\_r30 & 0.615 & 0.382 & -0.233 & 0.139 & \\
SoftImpute & C & popularity & f10\_loud\_r50 & 0.775 & 0.567 & -0.208 & 0.270 & \\
SoftImpute & C & popularity & f10\_weak\_r30 & 0.644 & 0.445 & -0.199 & 0.130 & \\
SoftImpute & C & popularity & f10\_weak\_r50 & 0.730 & 0.615 & -0.115 & 0.290 & \\
SoftImpute & C & popularity & f5\_loud\_r30 & 0.617 & 0.164 & -0.453 & 0.161 & \\
SoftImpute & C & popularity & f5\_loud\_r50 & 0.741 & 0.531 & -0.210 & 0.359 & \\
SoftImpute & C & popularity & f5\_weak\_r30 & 0.603 & 0.238 & -0.365 & 0.200 & \\
SoftImpute & C & popularity & f5\_weak\_r50 & 0.740 & 0.625 & -0.115 & 0.318 & \\
SoftImpute & S & cell & f10\_loud\_r30 & 0.593 & 0.544 & -0.049 & 0.160 & \\
SoftImpute & S & cell & f10\_loud\_r50 & 0.746 & 0.698 & -0.049 & 0.277 & \\
SoftImpute & S & cell & f10\_weak\_r30 & 0.582 & 0.665 & +0.083 & 0.193 & \\
SoftImpute & S & cell & f10\_weak\_r50 & 0.707 & 0.720 & +0.013 & 0.264 & \\
SoftImpute & S & cell & f5\_loud\_r30 & 0.606 & 0.557 & -0.049 & 0.170 & \\
SoftImpute & S & cell & f5\_loud\_r50 & 0.775 & 0.657 & -0.118 & 0.253 & \\
SoftImpute & S & cell & f5\_weak\_r30 & 0.637 & 0.555 & -0.082 & 0.167 & \\
SoftImpute & S & cell & f5\_weak\_r50 & 0.693 & 0.737 & +0.044 & 0.255 & \\
SoftImpute & S & mcar & f10\_loud\_r30 & 0.593 & 0.494 & -0.100 & 0.213 & \\
SoftImpute & S & mcar & f10\_loud\_r50 & 0.746 & 0.651 & -0.095 & 0.302 & \\
SoftImpute & S & mcar & f10\_weak\_r30 & 0.582 & 0.535 & -0.047 & 0.250 & \\
SoftImpute & S & mcar & f10\_weak\_r50 & 0.707 & 0.629 & -0.078 & 0.309 & \\
SoftImpute & S & mcar & f5\_loud\_r30 & 0.606 & 0.463 & -0.143 & 0.249 & \\
SoftImpute & S & mcar & f5\_loud\_r50 & 0.775 & 0.666 & -0.109 & 0.358 & \\
SoftImpute & S & mcar & f5\_weak\_r30 & 0.637 & 0.489 & -0.148 & 0.259 & \\
SoftImpute & S & mcar & f5\_weak\_r50 & 0.693 & 0.670 & -0.022 & 0.382 & \\
SoftImpute & S & popularity & f10\_loud\_r30 & 0.593 & 0.175 & -0.418 & 0.166 & \\
SoftImpute & S & popularity & f10\_loud\_r50 & 0.746 & 0.485 & -0.262 & 0.252 & \\
SoftImpute & S & popularity & f10\_weak\_r30 & 0.582 & 0.349 & -0.233 & 0.194 & \\
SoftImpute & S & popularity & f10\_weak\_r50 & 0.707 & 0.627 & -0.080 & 0.280 & \\
SoftImpute & S & popularity & f5\_loud\_r30 & 0.606 & 0.526 & -0.080 & 0.196 & \\
SoftImpute & S & popularity & f5\_loud\_r50 & 0.775 & 0.472 & -0.303 & 0.245 & \\
SoftImpute & S & popularity & f5\_weak\_r30 & 0.637 & 0.368 & -0.269 & 0.187 & \\
SoftImpute & S & popularity & f5\_weak\_r50 & 0.693 & 0.573 & -0.119 & 0.226 & \\
\end{longtable}}

\section{Factor analysis details}\label{factor-analysis-details}

\subsection{Estimator}\label{estimator}

We fit the exploratory factor analysis with the minimum-residual estimator,
applying a promax rotation whenever more than one factor is extracted and no
rotation otherwise. Where the default squared-multiple-correlation start for the
communalities errors on a singular correlation matrix, we retry the fit from a
unity diagonal. Only hard errors trigger that fallback, since benign warnings
still return a usable fit. The factor analysis used the R package \texttt{psych}
\citep{revelle2024}.

\subsection{Factor count}\label{factor-count}

We use Horn's parallel analysis \citep{horn1965} in its PC flavor. The observed
eigenvalues of the correlation matrix are compared position-by-position against
the 95th percentile of eigenvalues from 100 random \(n \times p\)
standard-normal matrices' correlation matrices, and
\[n_f = \#\{i : \lambda_i^{\text{obs}} > \lambda_i^{\text{cut}}\},\quad n_f \geq 2.\]
The count is then capped at 20, beyond which the bifactor fits become
prohibitively slow, and at \(\min(p-1,\, n-1,\, \operatorname{rank}(R) - 1)\),
since the completed and surrogate matrices are frequently rank-deficient or have
\(p \gg n\) and the estimator would otherwise error. If the fit still fails, we
decrement the count until it succeeds, and record the count actually used.

\subsection{Bifactor decomposition}\label{bifactor-decomposition}

The bifactor step uses the Schmid-Leiman
\citep{schmidleiman1957,reise2010,reise2012} transformation for EFA without sign
flipping. We record per cell the full Schmid-Leiman loading matrix (the general
factor plus the domain factors, per benchmark) and \(\omega_h\). From the
first-order solution we also record cumulative variance explained, per-factor
proportions, and the inter-factor correlation matrix \(\Phi\) with its mean
off-diagonal.

\section{Benchmark embedding}\label{benchmark-embedding}

Figure 1 places each benchmark in two dimensions using a composite distance
computed from the factor loadings rather than from the score matrix. This
section describes how that distance is built and how the figure is colored.

\subsection{Composite distancing}\label{composite-distancing}

Every bifactor solution assigns each benchmark a row of loadings, consisting of
its loading on the general factor followed by its loadings on each specific
factor. We take that row as the benchmark's vector within that solution. A
benchmark absent from a given solution does not contribute to it.

Within a single solution we compute the cosine distance between every pair of
benchmark vectors, which is one minus their cosine similarity, clipped to the
range 0 to 2. Cosine distance is invariant to the rotation and to the sign of
the factors, so per-solution distances remain comparable even though the factors
themselves are not. We then average each pair's distance across every solution
in which both benchmarks appear. This average is the composite distance, and it
is what the figure embeds.

The composite distance matrix is passed to UMAP \citep{mcinnes2018} as a
precomputed metric, with two components, ten neighbors, a minimum distance of
0.15, and a fixed random seed. Nothing is re-standardized at this stage, since
the distances already share a common scale.

\subsection{Uncomputable distance
handling}\label{uncomputable-distance-handling}

Two benchmarks that never appear together in any solution have no measured
distance between them. We fill each such entry with the mean of the distances
that one of the two benchmarks does have, falling back to the global mean when
neither has any. Luckily, missing entries are only observed in the R dataset
(which also has only one valid imputation, SoftImpute), with a very small
missing proportion. The R dataset has 318 benchmarks, and 160 out of 50,403
distances (0.32\%) were uncomputable. The statistics are identical for each
release year from 2020 to 2025 and for the aggregates. Uncomputable distances
were not observed in any other dataset, imputation, or release-year combination.

\subsection{Labels}\label{labels}

All benchmarks were labeled along three independent, non-exclusive axes: subject
(content domain), task (administration format), and language. A benchmark can
carry several subject tags, and some of those tags are nested within a broader
parent label (e.g.~medical under specialized\_domain). Labels were curated from
each benchmark's documentation, independent of the distance-geometry and
cohesion analyses.

\subsection{UMAP limitations}\label{umap-limitations}

UMAP preserves neither density nor global distance. Groups that appear tight or
far apart in two dimensions are partly an artifact of the embedding. We
therefore read the figure as a visual summary, and any claim we make about
clustering rests on the composite distance matrix itself.

\section{Release-date analysis}\label{release-date-analysis}

Here we check whether the factor structure changes with model release date, and
whether a benchmark's release date relates to its place in that structure. Both
analyses are exploratory and use the completed matrices of the main analysis, at
the end of the density ladder (10\%).

\subsection{Setup}\label{setup}

We group models by release year into four groups (2022 or earlier, 2023, 2024,
and 2025 or later), since the earlier years hold too few models to stand alone.
Release dates are known for 99.9\% of the models in each matrix. We take the
completed C and S matrices (standard collapse) of the imputers whose output rows
are the models themselves and that pass the quality filter (SoftImpute,
missForest and k-NN on C; SoftImpute and missForest on S), split their rows by
year group, and run each group through the same parallel analysis, EFA and
Schmid-Leiman pipeline as the pooled data
(\hyperref[factor-analysis-details]{Appendix~\ref*{factor-analysis-details}}).

A smaller sample moves \(\omega_h\) on its own. For each year group we therefore
also factor 50 random subsets of the same size, drawn from all dated models
regardless of year, and only count a group as different when its \(\omega_h\)
falls outside the 5th to 95th percentile of its subsets. We also compare the
general factor of each fit with the pooled one by Tucker's congruence.

It must be stressed, however, that each matrix is imputed once over all models
before the split. The missing cells of one year group are filled partly from the
other groups, and a benchmark that no model in the group took is filled entirely
from them. This happens often (\hyperref[tab:a4]{Table~\ref*{tab:a4}}), up to 42
of the 124 S benchmarks in the oldest group. Imputing each group on its own does
not work either, since a benchmark with no observation in a group cannot be
imputed from inside it. If the correlations between benchmarks are the same in
every year and only the level of performance shifts, the shared imputation does
no harm. If they are not, it pulls every group toward the pooled structure, and
the differences we report below are smaller than the true ones.

\begin{longtable}{@{}lrllll@{}}
\caption{Size and coverage of each release-year group. Each cell gives the number of models, the share of observed cells, and the number of benchmarks with no observed score in that group.}\label{tab:a4}\\
\toprule
Densifier & Benchmarks & $\leq 2022$ & 2023 & 2024 & $\geq 2025$ \\
\midrule
\endfirsthead
\caption[]{(continued)}\\
\toprule
Densifier & Benchmarks & $\leq 2022$ & 2023 & 2024 & $\geq 2025$ \\
\midrule
\endhead
\bottomrule
\endlastfoot
C & 78 & 87/20\%/10 & 196/14\%/6 & 294/12\%/21 & 92/12\%/10 \\
S & 124 & 86/14\%/42 & 196/9\%/30 & 291/9\%/26 & 93/11\%/16 \\
\end{longtable}

\subsection{Models by release year}\label{models-by-release-year}

\begin{figure}
\centering
\pandocbounded{\includegraphics[keepaspectratio]{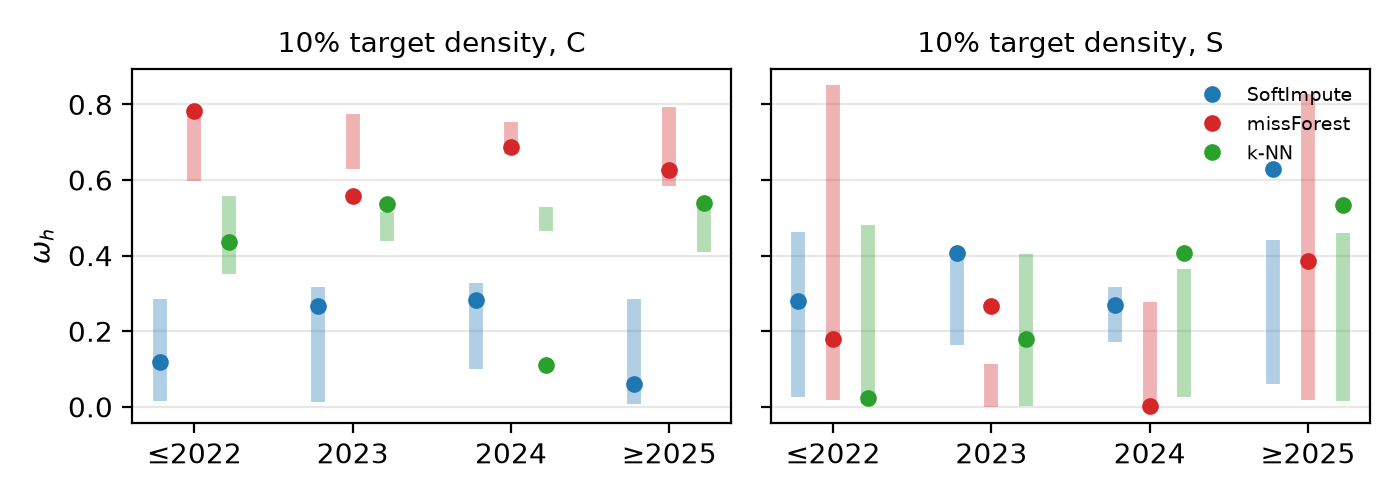}}
\caption{\(\omega_h\) of each release-year group (dots) against the 5th to 95th
percentile of 50 random subsets of the same size (bars), per imputer, on the C
(left) and S (right) matrices.\label{fig:release-year-omega}}
\end{figure}

The year-group \(\omega_h\) fits in this paragraph and in the figure come from
an earlier run of the pipeline (10\% density, before the final imputation chain)
and have not yet been repeated on the final matrices.
\hyperref[fig:release-year-omega]{Figure~\ref*{fig:release-year-omega}} plots
the \(\omega_h\) of each year group against the band of its random subsets.
There is no trend with release year. The imputers do not even agree on which
year has the strongest general factor. Still, the year groups differ from random
subsets more often than chance allows. Of the 24 fits, 8 fall outside the 90\%
band (6 above and 2 below) where 2 or 3 are expected, and the median congruence
of their general factor with the pooled one is 0.82, against 0.97 for the random
subsets. In other words, the structure does shift from one year to the next, but
in no consistent direction.

Newer models do score higher on the general factor
(\hyperref[tab:a5]{Table~\ref*{tab:a5}}). We score each model as the
loading-weighted sum of its standardized completed scores on the pooled general
factor, so the mean over all models is zero. In four of the five solutions the
mean score goes from between \(-1.1\) and \(-0.8\) for models of 2022 or earlier
to between \(0.4\) and \(1.2\) for models of 2025 or later. With SoftImpute on S
the scores are almost flat (\(0.01\) for 2022 or earlier, \(0.31\) for 2025 or
later), although the newest models still score highest. So newer models are
better on average. In the earlier run above, their general factor does not
explain more of the variance.

\begin{longtable}{@{}llrrrr@{}}
\caption{Mean general-factor score of each release-year group, with mean zero over all models.}\label{tab:a5}\\
\toprule
Densifier & Imputer & $\leq 2022$ & 2023 & 2024 & $\geq 2025$ \\
\midrule
\endfirsthead
\caption[]{(continued)}\\
\toprule
Densifier & Imputer & $\leq 2022$ & 2023 & 2024 & $\geq 2025$ \\
\midrule
\endhead
\bottomrule
\endlastfoot
C & SoftImpute & -1.12 & -0.43 & 0.51 & 0.37 \\
C & missForest & -1.07 & -0.49 & 0.32 & 1.03 \\
C & k-NN & -0.83 & -0.55 & 0.24 & 1.21 \\
S & SoftImpute & 0.01 & -0.16 & 0.01 & 0.31 \\
S & missForest & -1.05 & -0.24 & 0.26 & 0.69 \\
\end{longtable}

\subsection{Without imputation}\label{without-imputation}

Only six benchmarks are observed in every year group (BBH, GPQA, IFEval, MATH,
MMLU-Pro and MuSR, all from the Open LLM Leaderboard v2). For the 330 models
that have all six, we compute the correlation matrix of their observed scores
within each year group and take the share of variance of its first eigenvalue.
This share is 0.43 for 2022 or earlier, 0.71 for 2023, 0.72 for 2024 and 0.57
for 2025 or later. The oldest models sit near the floor of most of these
benchmarks (a mean MATH score of 1 out of 100, for instance), which lowers their
correlations, and the newest group holds only 20 models, with a bootstrap
interval ({[}0.50, 0.69{]}) that overlaps that of 2023 ({[}0.66, 0.81{]}). We
read no trend into this check either.

\subsection{Benchmark release date}\label{benchmark-release-date}

We also check whether the release year of a benchmark relates to its place in
the pooled solutions. Benchmark dates are less reliable than model dates
(\hyperref[release-dates]{Appendix~\ref*{release-dates}}), so we only use the
year. Since only the pooled loadings are needed here, all five solutions that
pass the quality filter on C and S enter
(\hyperref[tab:a6]{Table~\ref*{tab:a6}}).

First off, release year does not predict the general-factor loading. The
Spearman correlation between year and absolute loading is significant in only 1
of 5 solutions (missForest on S, \(\rho = -0.19\)).

Benchmarks from the same era do sit closer together in loading space. We group
benchmarks into four eras (2019 or earlier, 2020 to 2021, 2022 to 2023, and 2024
or later), and the mean within-era Euclidean distance between their loading
vectors is smaller than under 2,000 permutations of the era labels in all 5
solutions (\(z < -1.96\)).

Most of this closeness, however, comes from which models took each benchmark.
Newer benchmarks are mostly taken by newer models (the Spearman correlation
between a benchmark's year and the mean year of its models is 0.66 on S and 0.68
on C), and two benchmarks taken by the same models are filled in alike by the
imputers. We regress the rank of each pair's loading distance on the ranks of
its year gap and of the Jaccard distance between the two sets of observed
models. The co-observation term is the larger one in all 5 solutions (0.31 to
0.43, against \(-0.01\) to 0.13 for the year gap), although the year gap keeps a
significant effect of its own in 2 of 5 (permutation test with 500 relabelings).

The observed scores alone show a small era effect on both matrices. Among
benchmark pairs observed together on at least 30 models, pairs released four or
more years apart correlate at 0.40 on S and 0.46 on C, against 0.51 on both for
pairs released within a year of each other. Succinctly, most of the era effect
comes from the missingness pattern, and the small part left matches the
year-to-year shift we find for the models.

\begin{longtable}{@{}llrrlr@{}}
\caption{Benchmark release year against the pooled solutions. $\rho$ is the Spearman correlation between release year and absolute general-factor loading. Era $z$ compares within-era loading distance with permuted eras (negative means same-era benchmarks sit closer). The last two columns are standardized rank-regression coefficients of a pair's loading distance on its release-year gap (with permutation $p$) and on the Jaccard distance between its sets of observed models (co-obs.).}\label{tab:a6}\\
\toprule
Densifier & Imputer & $\rho$ & Era $z$ & Year gap ($p$) & Co-obs. \\
\midrule
\endfirsthead
\caption[]{(continued)}\\
\toprule
Densifier & Imputer & $\rho$ & Era $z$ & Year gap ($p$) & Co-obs. \\
\midrule
\endhead
\bottomrule
\endlastfoot
C & SoftImpute & 0.02 & -3.3 & -0.01 (0.618) & 0.37 \\*
C & missForest & 0.07 & -7.8 & 0.13 (0.014) & 0.43 \\*
C & k-NN & 0.05 & -4.5 & 0.08 (0.082) & 0.36 \\*
S & SoftImpute & 0.15 & -7.9 & 0.09 (0.000) & 0.31 \\*
S & missForest & -0.19 & -4.7 & 0.00 (0.494) & 0.37 \\*
\end{longtable}

\section{\texorpdfstring{Benchmark
\(G\)-rankings}{Benchmark G-rankings}}\label{benchmark-g-rankings}

\subsection{\texorpdfstring{Frequency and \(G\)-ranking
correlations}{Frequency and G-ranking correlations}}\label{frequency-and-g-ranking-correlations}

A pressing concern with regard to our findings on the top \(G\) benchmarks in
\hyperref[tab:g-rankings]{Table~\ref*{tab:g-rankings}} of the main content is to
what degree a benchmark's \(G\) factor loading is correlated with its
non-missing frequency. This is important to know, as our datasets can possess
higher or lower correlations introduced as artifacts by our imputation methods.

Naively, as shown in
\hyperref[tab:g-rank-freq-corr]{Table~\ref*{tab:g-rank-freq-corr}} below,
benchmark frequency \emph{is} correlated with \(G\) loadings. Unsigned averages
suggest this correlation is modest (\(r = 0.2740\)), but the correlations are
quite dispersed, and some dataset-imputation combinations can be as high as
\textasciitilde0.5. Thus, identifying benchmarks which proxy a supposed latent
\(G\) factor well requires adjusting the statistics with respect to their
frequency.

\begin{longtable}{@{}llrrr@{}}
\caption{Correlations between benchmark frequency and their $G$ loadings. $k$ = number of factors extracted. Only accepted imputations with $wR^2 > 0.4$ are shown.}\label{tab:g-rank-freq-corr}\\
\toprule
Method & Dataset & $k$ & $r$ & $N$ \\
\midrule
\endfirsthead
\caption[]{(continued)}\\
\toprule
Method & Dataset & $k$ & $r$ & $N$ \\
\midrule
\endhead
\bottomrule
\endlastfoot
\multicolumn{5}{@{}l}{[$k = 2$ rows removed]} \\
k-NN & C Std. & 8 & +0.3447 & 78 \\
missForest & C Std. & 4 & +0.3200 & 78 \\
missForest & S Std. & 5 & +0.2845 & 124 \\
SoftImpute & C Std. & 10 & +0.4880 & 78 \\
SoftImpute & R Aggr. & 20 & -0.1257 & 310 \\
SoftImpute & S Std. & 8 & -0.0813 & 124 \\
\textbf{Average} & & & \textbf{+0.2050} & 6 (groups) \\
\textbf{Average $\lvert r\lvert$} & & & \textbf{+0.2740} & \\
\end{longtable}

\subsection{Frequency adjustment}\label{frequency-adjustment}

Let \(C\) denote a factor cell: a method and dataset combination for which
factor loadings are available. Within each cell \(C\), benchmarks are ranked by
absolute loading \(|g_i|\) (\(i\) indexes benchmarks). The rank \(r_i\) is
normalized within the cell,

\[a_i = \frac{r_i - 1}{n - 1}, \qquad r_i \in \{1, \dots, n\},\]

so \(a_i = 0\) at the top of the cell and \(a_i = 1\) at the bottom. This makes
cells with different numbers of benchmarks \(n\) comparable. Each benchmark's
raw score is the mean of its \(a_i\) over the cells in which it appears.
Benchmarks appearing in fewer than two cells are excluded.

For a dataset with \(m\) models, the frequency of benchmark \(i\) is the
proportion of models with an observed score,

\[f_i = \frac{\#\{\text{models with non-missing score for } i\}}{m}.\]

Columns with fewer than two observations or zero variance are excluded before
computing \(f_i\). Where several datasets contribute, the benchmark's frequency
is the mean of its per-dataset \(f_i\).

As shown in
\hyperref[tab:g-rank-cellwise-corr]{Table~\ref*{tab:g-rank-cellwise-corr}}
below, within most cells, frequently measured benchmarks obtain higher \(|g_i|\)
and lower (better) \(a_i\). A benchmark may therefore rank highly partly because
it is measured often rather than because it is a strong indicator of \(G\).

The adjustment is applied within each cell rather than to the pooled averages.
Within cell \(C\), the normalized ranks are regressed on the cell's own
frequencies by ordinary least squares,

\[a_i = \alpha + \beta f_i + \varepsilon_i, \qquad
\hat{\varepsilon}_i = (a_i - \bar{a}) - \hat{\beta}\,(f_i - \bar{f}),\]

and the residual \(\hat{\varepsilon}_i\) replaces \(a_i\). By construction the
residuals are uncorrelated with \(f_i\) within the cell, so \(\hat{\beta}\)
removes exactly the linear component of the within-cell rank--frequency
association. A cell with fewer than three frequency--rank pairs, or with zero
frequency variance, cannot be fitted. Its benchmarks receive centered raw ranks
\((a_i - \bar{a})\), and such cells are counted and reported.

The adjusted score of benchmark \(i\) is the mean of its residuals over the
cells in which it appears. Rankings are reported on this scale, with the raw
average rank retained for comparison.

\subsection{Diagnostics}\label{diagnostics}

We present two diagnostics that justify the method of adjustment.
\hyperref[tab:g-rank-cellwise-corr]{Table~\ref*{tab:g-rank-cellwise-corr}} below
reports, within each cell, the Pearson correlation \(r(f_i, a_i)\), with the
mean \(r\), the mean \(|r|\), and the counts of negative and positive cells.
Negative \(r\) indicates that more frequently measured benchmarks load higher on
\(G\).

\begin{longtable}{@{}llrr@{}}
\caption{Correlations between benchmark frequency and their $G$ rankings. Mean $r = -0.151$, mean $\lvert r \rvert = 0.306$; 4 negative and 2 positive cells.}\label{tab:g-rank-cellwise-corr}\\
\toprule
Method & Dataset & $N$ & $r$ \\
\midrule
\endfirsthead
\caption[]{(continued)}\\
\toprule
Method & Dataset & $N$ & $r$ \\
\midrule
\endhead
\bottomrule
\endlastfoot
k-NN & C Std. & 78 & -0.413 \\
missForest & C Std. & 78 & -0.234 \\
missForest & S Std. & 124 & -0.318 \\
SoftImpute & C Std. & 78 & -0.406 \\
SoftImpute & R Aggr. & 310 & +0.072 \\
SoftImpute & S Std. & 124 & +0.394 \\
\end{longtable}

Second, \hyperref[tab:g-rank-pooled-corr]{Table~\ref*{tab:g-rank-pooled-corr}}
reports the pooled correlation between mean frequency and mean rank before and
after the adjustment, stratified by the number of cells \(k\) in which a
benchmark appears. The within-cell association does not survive pooling with a
consistent sign: cells disagree in direction, so the pooled raw correlation is
small even though the within-cell correlations are not.

\begin{longtable}{@{}lrrr@{}}
\caption{Pooled correlation between mean frequency and mean rank.}\label{tab:g-rank-pooled-corr}\\
\toprule
$N$ cells & $N$ benchmarks & $r$ raw & $r$ adjusted \\
\midrule
\endfirsthead
\caption[]{(continued)}\\
\toprule
$N$ cells & $N$ benchmarks & $r$ raw & $r$ adjusted \\
\midrule
\endhead
\bottomrule
\endlastfoot
3 & 46 & -0.290 & -0.298 \\
6 & 78 & -0.266 & +0.113 \\
Pooled & 124 & -0.135 & +0.095 \\
\end{longtable}

For this reason, we report the cellwise-frequency-adjusted average normalized
rank as the primary ordering of benchmarks
(\hyperref[tab:g-rankings]{Table~\ref*{tab:g-rankings}} in the main text). This
ordering removes the linear component of the within-cell rank--frequency
association, which reaches \(|r| \approx 0.3\) in the average cell, before
averaging across cells. The raw average normalized rank is retained alongside it
for comparison, as the two orderings agree closely. The adjustment mainly
matters for benchmarks measured across many cells, where the within-cell
association is strongest and where the adjusted ranking reorders the raw
ranking.

The full ranking of all 124 benchmarks is given in
\hyperref[full-g-rankings]{Appendix~\ref*{full-g-rankings}}.

\section{General factor and model size}\label{general-factor-and-model-size}

We compute each model's general-factor score from the loadings and compare it
with the parameter count within model families that have at least three distinct
sizes. R Aggr. is omitted because its rows merge models of different sizes.

\begin{longtable}{@{}llrrrrr@{}}
\caption{Within-family association between general-factor score and log parameter count. $r$ is the Pearson correlation after subtracting family means from both variables; the last column counts families with a positive Spearman correlation.}\label{tab:g-model-size}\\
\toprule
Method & Dataset & Families & Models & $r$ & $p$ & Families $\rho > 0$ \\
\midrule
\endfirsthead
\caption[]{(continued)}\\
\toprule
Method & Dataset & Families & Models & $r$ & $p$ & Families $\rho > 0$ \\
\midrule
\endhead
\bottomrule
\endlastfoot
k-NN & C Std. & 35 & 142 & +0.651 & $<0.001$ & 32/35 \\
missForest & S Std. & 35 & 141 & +0.591 & $<0.001$ & 30/35 \\
missForest & C Std. & 35 & 142 & +0.499 & $<0.001$ & 24/35 \\
SoftImpute & C Std. & 35 & 142 & +0.359 & $<0.001$ & 26/35 \\
SoftImpute & S Std. & 35 & 141 & -0.107 & 0.2 & 14/35 \\
\end{longtable}

\section{Label cohesion results}\label{label-cohesion-results}

We present the full label cohesion results for all labels here. From the subject
category, only \texttt{finance} and \texttt{professional\_writing} show
significantly strong cohesion, both of which are very specific labels with low
median n.~The next few labels show moderate to weak cohesion at best, a lot of
which do not have enough significance to distinguish from chance. Labeling by
language surprisingly shows many weakly cohesive structures, especially for
\texttt{monolingual\_non\_english}, which is an aggregate label. Individual
labels for each language are not used due to low n.~All of the results support
our claim that benchmark clusters are weakly cohesive. Across all 57 labels, the
median of the per-label median \(A\) is \(+0.094\). Within a single solution and
before FDR correction, 16 to 49\% of the scorable labels reach \(p < 0.05\),
depending on the solution and on whether the \(G\) loadings are included in the
loading vectors.

\begin{longtable}{@{}lrcrr@{}}
\caption{Cohesion by task-format label, sorted by descending median A.}\label{tab:cohesion-task-type}\\
\toprule
Label & Median A & A across cells & Significant & Median n \\
\midrule
\endfirsthead
\caption[]{(continued)}\\
\toprule
Label & Median A & A across cells & Significant & Median n \\
\midrule
\endhead
\bottomrule
\endlastfoot
\texttt{likelihood\_probe} & +0.768 & +0.181 to +0.879 & 3/5 & 4 \\
\texttt{extraction} & +0.459 & +0.026 to +0.772 & 3/5 & 5 \\
\texttt{interactive} & +0.244 & +0.159 to +0.587 & 2/5 & 6 \\
\texttt{short\_qa} & +0.198 & +0.060 to +0.462 & 2/9 & 15 \\
\texttt{classification} & +0.186 & +0.008 to +0.285 & 0/9 & 15 \\
\texttt{multiple\_choice} & +0.079 & -0.003 to +0.235 & 1/9 & 36 \\
\texttt{conversation} & +0.026 & +0.025 to +0.028 & 0/2 & 7 \\
\texttt{long\_reasoning} & +0.014 & -0.211 to +0.370 & 0/9 & 9 \\
\texttt{sentence\_completion} & -0.035 & -0.646 to +0.128 & 0/9 & 5 \\
\texttt{long\_form\_generation} & -0.036 & -0.144 to +0.037 & 0/9 & 32 \\
\end{longtable}

\begin{longtable}{@{}lrcrr@{}}
\caption{Cohesion by language-coverage label, sorted by descending median A.}\label{tab:cohesion-language}\\
\toprule
Label & Median A & A across cells & Significant & Median n \\
\midrule
\endfirsthead
\caption[]{(continued)}\\
\toprule
Label & Median A & A across cells & Significant & Median n \\
\midrule
\endhead
\bottomrule
\endlastfoot
\texttt{monolingual\_non\_english} & +0.272 & +0.024 to +0.771 & 6/9 & 19 \\
\texttt{multilingual} & +0.183 & -0.006 to +0.435 & 2/5 & 9 \\
\texttt{crosslingual} & +0.135 & +0.083 to +0.193 & 0/5 & 10 \\
\texttt{english} & -0.009 & -0.176 to +0.004 & 0/9 & 85 \\
\end{longtable}

\begin{longtable}{@{}lrcrr@{}}
\caption{Cohesion by subject-matter label, sorted by descending median A.}\label{tab:cohesion-subject}\\
\toprule
Label & Median A & A across cells & Significant & Median n \\
\midrule
\endfirsthead
\caption[]{(continued)}\\
\toprule
Label & Median A & A across cells & Significant & Median n \\
\midrule
\endhead
\bottomrule
\endlastfoot
\texttt{finance} & +0.910 & +0.814 to +0.912 & 5/5 & 4 \\
\texttt{professional\_writing} & +0.612 & +0.610 to +0.613 & 1/2 & 5 \\
\texttt{factuality} & +0.485 & +0.132 to +0.577 & 2/5 & 7 \\
\texttt{world\_knowledge} & +0.399 & +0.119 to +0.602 & 2/9 & 7 \\
\texttt{creativity} & +0.378 & +0.377 to +0.378 & 0/2 & 4 \\
\texttt{reading\_comprehension} & +0.331 & -0.025 to +0.485 & 3/9 & 9 \\
\texttt{commonsense} & +0.329 & +0.189 to +0.514 & 2/9 & 7 \\
\texttt{sentiment} & +0.319 & -0.071 to +0.733 & 2/5 & 5 \\
\texttt{code} & +0.278 & +0.196 to +0.589 & 4/9 & 10 \\
\texttt{agentic} & +0.257 & +0.199 to +0.542 & 0/5 & 5 \\
\texttt{language\_understanding} & +0.253 & +0.251 to +0.254 & 0/2 & 10 \\
\texttt{science} & +0.233 & -0.014 to +0.614 & 1/9 & 11 \\
\texttt{language\_modelling} & +0.205 & +0.205 to +0.206 & 0/2 & 5 \\
\texttt{translation} & +0.190 & +0.004 to +0.221 & 0/5 & 9 \\
\texttt{linguistic\_competence} & +0.180 & -0.142 to +0.481 & 0/5 & 4 \\
\texttt{encyclopedic} & +0.165 & +0.078 to +0.574 & 2/9 & 23 \\
\texttt{social\_media} & +0.153 & +0.056 to +0.477 & 0/5 & 6 \\
\texttt{medical} & +0.139 & +0.139 to +0.139 & 2/2 & 40 \\
\texttt{games} & +0.113 & +0.112 to +0.113 & 0/2 & 6 \\
\texttt{structured\_data} & +0.110 & +0.109 to +0.111 & 0/2 & 15 \\
\texttt{recall} & +0.094 & -0.001 to +0.617 & 4/9 & 21 \\
\texttt{cognitive} & +0.086 & +0.086 to +0.087 & 0/2 & 12 \\
\texttt{dialogue} & +0.066 & +0.062 to +0.070 & 0/2 & 5 \\
\texttt{society\_culture} & +0.047 & +0.045 to +0.049 & 0/2 & 13 \\
\texttt{multi\_subject} & +0.037 & +0.011 to +0.320 & 0/9 & 28 \\
\texttt{generation} & +0.030 & -0.185 to +0.134 & 0/5 & 11 \\
\texttt{language\_processing} & +0.029 & -0.199 to +0.366 & 1/9 & 26 \\
\texttt{math} & +0.028 & -0.152 to +0.366 & 0/9 & 10 \\
\texttt{specialized\_domain} & +0.013 & -0.159 to +0.175 & 0/9 & 13 \\
\texttt{text\_classification} & -0.002 & -0.002 to -0.002 & 0/2 & 12 \\
\texttt{source\_genre} & -0.002 & -0.155 to +0.137 & 1/9 & 39 \\
\texttt{reasoning} & -0.017 & -0.071 to +0.008 & 0/9 & 19 \\
\texttt{logical\_reasoning} & -0.033 & -0.337 to +0.071 & 0/9 & 6 \\
\texttt{summarization} & -0.035 & -0.035 to -0.034 & 0/2 & 13 \\
\texttt{instruction\_following} & -0.045 & -0.045 to -0.044 & 0/2 & 19 \\
\texttt{alignment} & -0.051 & -0.336 to +0.092 & 1/9 & 26 \\
\texttt{natural\_language\_inference} & -0.052 & -0.054 to -0.051 & 0/2 & 6 \\
\texttt{toxicity} & -0.070 & -0.317 to +0.224 & 0/5 & 6 \\
\texttt{fiction} & -0.087 & -0.088 to -0.086 & 0/2 & 4 \\
\texttt{safety} & -0.094 & -0.901 to +0.236 & 1/9 & 8 \\
\texttt{retrieval} & -0.115 & -0.118 to -0.113 & 0/2 & 6 \\
\texttt{bias\_fairness} & -0.222 & -0.228 to -0.216 & 0/2 & 4 \\
\texttt{news} & -0.222 & -0.727 to +0.188 & 0/9 & 5 \\
\end{longtable}

\section{Imputation Results}\label{imputation-results}

The table below lists the weighted held-out \(R^2\) (\(wR^2\)) of every
dataset-imputer combination. Rows are sorted by \(wR^2\). As described in the
results, only 6 of the 18 combinations pass the \(wR^2 \geq 0.4\) quality
filter.

\begin{longtable}{@{}llr@{}}
\caption{Results of all imputation runs, sorted by $wR^2$.}\label{tab:imputation-results-all}\\
\toprule
Dataset & Imputer & $wR^2$ \\
\midrule
\endfirsthead
\caption[]{(continued)}\\
\toprule
Dataset & Imputer & $wR^2$ \\
\midrule
\endhead
\bottomrule
\endlastfoot
C Std. & SoftImpute & 0.586 \\
C Std. & missForest & 0.566 \\
S Std. & missForest & 0.455 \\
C Std. & k-NN & 0.442 \\
R Aggr. & SoftImpute & 0.441 \\
S Std. & SoftImpute & 0.428 \\
C Aggr. & SoftImpute & 0.398 \\
R Std. & SoftImpute & 0.355 \\
S Std. & k-NN & 0.326 \\
C Aggr. & missForest & 0.263 \\
S Aggr. & SoftImpute & 0.262 \\
C Aggr. & k-NN & 0.170 \\
S Aggr. & k-NN & 0.016 \\
R Aggr. & k-NN & 0.004 \\
S Aggr. & missForest & 0.001 \\
R Std. & missForest & 0.000 \\
R Aggr. & missForest & -0.001 \\
R Std. & k-NN & -0.004 \\
\end{longtable}

\section{Omega sensitivity}\label{omega-sensitivity}

To quantify to what degree missing observations affect our results, we run
leave-one-covariate-out (LOCO) factor analyses for each valid dataset. For each
benchmark in the dataset, we run factor analysis with the benchmark left out,
and store the difference in \(\omega_h\) as a measure of sensitivity.
\hyperref[tab:omega-sensitivity]{Table~\ref*{tab:omega-sensitivity}} shows the
correlations between benchmark frequency (normalized within the dataset) and the
deltas. Signed averages of \(r\) show a smaller correlation at \(r=0.157\), but
this may simply be because signed correlations cancel out to 0. Average of
unsigned, absolute \(r\) yielded a larger but still modest correlation of
\(r=0.241\).

\begin{longtable}{@{}llrrr@{}}
\caption{Correlations between benchmark observation and their $\Delta\omega_h$. $k$ = number of factors extracted. Only accepted imputations with $wR^2 > 0.4$ are shown.}\label{tab:omega-sensitivity}\\
\toprule
Method & Dataset & $k$ & $r$ & $N$ \\
\midrule
\endfirsthead
\caption[]{(continued)}\\
\toprule
Method & Dataset & $k$ & $r$ & $N$ \\
\midrule
\endhead
\bottomrule
\endlastfoot
\multicolumn{5}{@{}l}{[$k = 2$ rows removed]} \\
k-NN & C Std. & 8 & +0.3251 & 78 \\
missForest & C Std. & 4 & -0.0222 & 78 \\
missForest & S Std. & 5 & +0.4566 & 124 \\
SoftImpute & C Std. & 10 & +0.3287 & 78 \\
SoftImpute & R Aggr. & 20 & -0.2289 & 310 \\
SoftImpute & S Std. & 8 & +0.0841 & 124 \\
\textbf{Average $r$} & & & \textbf{+0.1572} & 6 (groups) \\
\textbf{Average $\lvert r \lvert$} & & & \textbf{+0.2409} & \\
\end{longtable}

\clearpage

\section{\texorpdfstring{Full
\(G\)-rankings}{Full G-rankings}}\label{full-g-rankings}

\hyperref[tab:full-g-rankings]{Table~\ref*{tab:full-g-rankings}} lists all 124
benchmarks used across our analyses, sorted by their rank order as in
\hyperref[tab:g-rankings]{Table~\ref*{tab:g-rankings}} of the main text. The
ranking method and its diagnostics are described in
\hyperref[benchmark-g-rankings]{Appendix~\ref*{benchmark-g-rankings}}.

\begingroup\normalsize\setlength{\tabcolsep}{3pt}
\begin{longtable}{@{}r>{\raggedright\arraybackslash}p{0.21\textwidth}rrlrrr>{\raggedright\arraybackslash}p{0.25\textwidth}@{}}
\caption{All used 124 benchmarks, sorted by their average normalized rank order (ANR) of their $G$ factor loadings, residualized (RANR) against their frequency. The normalized rank order ranges from 0 to 1. 0 = ranked first, 1 = ranked last. $N$ cells = number of EFA solutions with that benchmark. CI and Best/Worst refers to ANR. Leading zeros are omitted. Reference is the paper introducing the benchmark; $^\dagger$ marks benchmarks without one, cited by the source of their scores.}\label{tab:full-g-rankings}\\
\toprule
No & Benchmark & $\rho_\epsilon$ & $\rho$ & 95\% CI & Best & Worst & $N$ & Reference \\
\midrule
\endfirsthead
\caption[]{(continued)}\\
\toprule
No & Benchmark & $\rho_\epsilon$ & $\rho$ & 95\% CI & Best & Worst & $N$ & Reference \\
\midrule
\endhead
\bottomrule
\endlastfoot
1 & simpleqa & -.312 & .186 & [-.199, .572] & .041 & .350 & 3 & \citealp{wei2024measuring} \\
2 & kaggle\_yulongt\_facts\_parametric & -.307 & .189 & [-.491, .870] & .006 & .504 & 3 & \citealp{kaggle_benchmarks}$^\dagger$ \\
3 & kaggle\_andrewmingwang\_asset\_ops\_bench & -.267 & .228 & [-.424, .879] & .049 & .528 & 3 & \citealp{constantinides2025failuresensoriq} \\
4 & kaggle\_olamysiak\_eclektic & -.242 & .256 & [.049, .463] & .175 & .341 & 3 & \citealp{goldman2025eclektic} \\
5 & tydiqa & -.235 & .260 & [.046, .474] & .195 & .358 & 3 & \citealp{clark2020tydi} \\
6 & facts\_search & -.232 & .262 & [-.208, .732] & .104 & .472 & 3 & \citealp{cheng2025facts} \\
7 & medqa & -.192 & .316 & [.016, .615] & .065 & .819 & 6 & \citealp{jin2020what} \\
8 & pwc\_arc\_challenge & -.186 & .351 & [.083, .619] & .104 & .767 & 6 & \citealp{clark2018think} \\
9 & thai\_exam\_a\_level & -.175 & .354 & [.020, .688] & .123 & .959 & 6 & \citealp{pipatanakul2023typhoon} \\
10 & bbq & -.162 & .349 & [-.013, .711] & .013 & .967 & 6 & \citealp{parrish2021bbq} \\
11 & legalbench & -.157 & .353 & [.079, .627] & .078 & .748 & 6 & \citealp{guha2023legalbench} \\
12 & pwc\_piqa & -.156 & .381 & [.053, .710] & .033 & .939 & 6 & \citealp{bisk2019piqa} \\
13 & thai\_exam\_tpat1 & -.156 & .373 & [.013, .733] & .000 & .878 & 6 & \citealp{pipatanakul2023typhoon} \\
14 & openbookqa & -.154 & .343 & [-.021, .708] & .026 & .862 & 6 & \citealp{mihaylov2018can} \\
15 & thaiexam & -.153 & .376 & [-.007, .758] & .074 & .984 & 6 & \citealp{pipatanakul2023typhoon} \\
16 & simple\_safety\_tests & -.149 & .363 & [.029, .697] & .039 & .886 & 6 & \citealp{vidgen2023simplesafetytests} \\
17 & anthropic\_red\_team & -.140 & .372 & [.023, .721] & .026 & .902 & 6 & \citealp{ganguli2022red} \\
18 & wisesight & -.136 & .358 & [-.495, 1.212] & .073 & .740 & 3 & \citealp{susanto2025sea}$^\dagger$ \\
19 & wmt\_14 & -.136 & .374 & [.018, .731] & .114 & 1.000 & 6 & \citealp{bojar2014findings} \\
20 & chatbot\_arena & -.134 & .375 & [.116, .633] & .026 & .650 & 6 & \citealp{zheng2023judging} \\
21 & kaggle\_andrewmingwang\_scicode\_subproblem\_with\_background & -.132 & .396 & [.258, .533] & .179 & .519 & 6 & \citealp{tian2024scicode} \\
22 & kaggle\_andrewmingwang\_scicode\_main\_with\_background & -.124 & .374 & [.149, .600] & .276 & .455 & 3 & \citealp{tian2024scicode} \\
23 & xquad & -.119 & .376 & [.190, .562] & .298 & .447 & 3 & \citealp{artetxe2019cross} \\
24 & blimp & -.119 & .376 & [-.758, 1.510] & .087 & .902 & 3 & \citealp{warstadt2019blimp} \\
25 & xstest & -.116 & .396 & [.137, .654] & .182 & .821 & 6 & \citealp{rottger2023xstest} \\
26 & synthetic\_reasoning & -.113 & .409 & [.082, .736] & .000 & .809 & 6 & \citealp{helm2023} \\
27 & synthetic\_reasoning\_natural & -.112 & .410 & [.093, .728] & .041 & .870 & 6 & \citealp{helm2023} \\
28 & nusax & -.111 & .384 & [-.361, 1.130] & .130 & .715 & 3 & \citealp{winata2022nusax} \\
29 & truthfulqa & -.110 & .322 & [.140, .504] & .156 & .642 & 6 & \citealp{lin2021truthfulqa} \\
30 & mmlu\_prox & -.107 & .420 & [.118, .722] & .152 & .805 & 6 & \citealp{xuan2025mmlu} \\
31 & ice & -.104 & .390 & [-.399, 1.179] & .024 & .593 & 3 & \citealp{helm2023}$^\dagger$ \\
32 & xcopa & -.101 & .394 & [-.474, 1.261] & .181 & .797 & 3 & \citealp{ponti2020xcopa} \\
33 & twitter\_aae & -.099 & .395 & [-.686, 1.477] & .000 & .862 & 3 & \citealp{blodgett2016demographic} \\
34 & real\_toxicity\_prompts & -.099 & .431 & [.125, .738] & .097 & .870 & 6 & \citealp{gehman2020realtoxicityprompts} \\
35 & lindsea\_pragmatics\_scalar\_implicatures\_id & -.097 & .397 & [-.050, .845] & .217 & .577 & 3 & \citealp{leong2023bhasa} \\
36 & kaggle\_andrewmingwang\_simpleqa\_verified & -.093 & .438 & [.153, .723] & .049 & .727 & 6 & \citealp{haas2025simpleqa} \\
37 & bfcl & -.090 & .441 & [.118, .764] & .071 & .846 & 6 & \citealp{bfcl_leaderboard} \\
38 & arena\_hard\_auto & -.084 & .447 & [.036, .858] & .013 & .903 & 6 & \citealp{li2024from} \\
39 & kaggle\_andrewmingwang\_scicode\_subproblem\_standard & -.081 & .447 & [.251, .643] & .211 & .779 & 6 & \citealp{tian2024scicode} \\
40 & msmarco\_trec & -.080 & .414 & [-.404, 1.232] & .081 & .740 & 3 & \citealp{campos2016ms} \\
41 & lindsea\_syntax\_minimal\_pairs\_id & -.077 & .418 & [-.683, 1.518] & .122 & .927 & 3 & \citealp{leong2023bhasa} \\
42 & indicqa & -.076 & .419 & [-.153, .991] & .187 & .647 & 3 & \citealp{doddapaneni2022leaving} \\
43 & bbh & -.073 & .317 & [-.003, .638] & .052 & .780 & 6 & \citealp{suzgun2022challenging} \\
44 & uitvsfc & -.072 & .423 & [.039, .808] & .325 & .602 & 3 & \citealp{nguyen2018uit} \\
45 & flores\_en\_vi & -.071 & .424 & [.090, .757] & .293 & .561 & 3 & \citealp{costajussa2022no} \\
46 & mmlu & -.070 & .367 & [.027, .706] & .039 & .829 & 6 & \citealp{hendrycks2021} \\
47 & indicsentiment & -.068 & .427 & [-.498, 1.351] & .171 & .854 & 3 & \citealp{doddapaneni2022leaving} \\
48 & mmlu\_pro & -.065 & .325 & [-.019, .668] & .000 & .797 & 6 & \citealp{wang2024mmlu} \\
49 & kaggle\_andrewmingwang\_facts & -.064 & .430 & [-.157, 1.017] & .163 & .610 & 3 & \citealp{jacovi2025facts} \\
50 & kaggle\_vijitsingh1\_mgsm\_english & -.060 & .468 & [.268, .667] & .171 & .756 & 6 & \citealp{shi2022language} \\
51 & arabicmmlu & -.058 & .473 & [.120, .826] & .019 & .857 & 6 & \citealp{koto2024arabicmmlu} \\
52 & madinah\_qa & -.057 & .474 & [.087, .860] & .003 & .883 & 6 & \citealp{helm_arabic}$^\dagger$ \\
53 & kaggle\_sripalthilakraj\_global\_mmlu\_lite\_english & -.054 & .443 & [-.330, 1.216] & .094 & .691 & 3 & \citealp{singh2024global} \\
54 & msmarco\_regular & -.051 & .443 & [-.456, 1.343] & .089 & .813 & 3 & \citealp{campos2016ms} \\
55 & flores\_th\_en & -.048 & .447 & [-.023, .917] & .244 & .618 & 3 & \citealp{costajussa2022no} \\
56 & multiloko & -.045 & .480 & [.124, .835] & .029 & .909 & 6 & \citealp{hupkes2025multiloko} \\
57 & cogbench & -.043 & .453 & [-.590, 1.495] & .106 & .919 & 3 & \citealp{codaforno2024cogbench} \\
58 & gsm & -.040 & .449 & [.022, .877] & .000 & .935 & 6 & \citealp{cobbe2021training} \\
59 & artificial\_analysis\_intelligence & -.033 & .505 & [.223, .786] & .061 & .818 & 6 & \citealp{artificialanalysis_index} \\
60 & legal\_support & -.030 & .492 & [.145, .838] & .091 & .951 & 6 & \citealp{helm2023} \\
61 & mbpp & -.030 & .510 & [.093, .927] & .039 & .976 & 6 & \citealp{austin2021program} \\
62 & xnli & -.027 & .468 & [-.256, 1.192] & .220 & .789 & 3 & \citealp{conneau2018xnli} \\
63 & thai\_exam\_tgat & -.026 & .503 & [.068, .937] & .023 & .992 & 6 & \citealp{pipatanakul2023typhoon} \\
64 & aratrust & -.025 & .505 & [.091, .919] & .016 & .961 & 6 & \citealp{alghamdi2024aratrust} \\
65 & math\_chain\_of\_thought & -.021 & .468 & [.014, .923] & .013 & .997 & 6 & \citealp{hendrycks2021measuring} \\
66 & kaggle\_andrewmingwang\_scicode\_main\_standard & -.010 & .486 & [-.031, 1.004] & .268 & .683 & 3 & \citealp{tian2024scicode} \\
67 & quac & -.009 & .514 & [.248, .780] & .024 & .753 & 6 & \citealp{choi2018quac} \\
68 & summarization\_xsum & -.003 & .527 & [.270, .784] & .138 & .831 & 6 & \citealp{narayan2018dont} \\
69 & arc & -.003 & .435 & [.161, .709] & .073 & .878 & 6 & \citealp{clark2018think} \\
70 & gsm8k & -.002 & .423 & [.152, .695] & .117 & .841 & 6 & \citealp{cobbe2021training} \\
71 & arabic\_exams & -.001 & .530 & [.116, .943] & .033 & .883 & 6 & \citealp{hardalov2020exams} \\
72 & alghafa & .012 & .542 & [.112, .973] & .008 & .896 & 6 & \citealp{almazrouei2023alghafa} \\
73 & babi\_qa & .017 & .538 & [.224, .853] & .008 & .857 & 6 & \citealp{weston2015ai} \\
74 & lindsea\_pragmatics\_presuppositions\_id & .018 & .513 & [-.002, 1.028] & .294 & .707 & 3 & \citealp{leong2023bhasa} \\
75 & gpqa\_diamond & .019 & .548 & [.353, .743] & .340 & .870 & 6 & \citealp{rein2023gpqa} \\
76 & kaggle\_sjmikler\_livecodebench & .019 & .546 & [.304, .788] & .236 & .883 & 6 & \citealp{jain2024livecodebench} \\
77 & pwc\_arc\_easy & .022 & .518 & [-.561, 1.597] & .016 & .773 & 3 & \citealp{clark2018think} \\
78 & kaggle\_aminmohamed\-mohami\_browsecomp & .022 & .551 & [.221, .882] & .065 & .935 & 6 & \citealp{wei2025browsecomp} \\
79 & disinformation\_reiteration & .023 & .553 & [.249, .857] & .179 & .896 & 6 & \citealp{helm2023} \\
80 & natural\_qa\_closedbook & .027 & .518 & [.186, .850] & .154 & .877 & 6 & \citealp{kwiatkowski2019natural} \\
81 & alrage & .031 & .562 & [.186, .937] & .089 & .974 & 6 & \citealp{helm_arabic}$^\dagger$ \\
82 & ifeval & .032 & .421 & [.254, .589] & .208 & .660 & 6 & \citealp{zhou2023instruction} \\
83 & harmbench & .032 & .544 & [.314, .774] & .165 & .748 & 6 & \citealp{mazeika2024harmbench} \\
84 & wikifact & .035 & .557 & [.261, .854] & .049 & .828 & 6 & \citealp{helm2023} \\
85 & swe\_bench & .043 & .535 & [-.512, 1.581] & .084 & .919 & 3 & \citealp{jimenez2023swe} \\
86 & disinformation\_wedging & .046 & .577 & [.320, .833] & .154 & .829 & 6 & \citealp{helm2023} \\
87 & gpqa & .048 & .438 & [.088, .787] & .078 & .976 & 6 & \citealp{rein2023gpqa} \\
88 & aime25 & .052 & .580 & [.324, .837] & .301 & .961 & 6 & \citealp{maa_aime} \\
89 & vihsd & .058 & .553 & [.319, .786] & .447 & .626 & 3 & \citealp{luu2021large} \\
90 & math\_regular & .058 & .598 & [.202, .993] & .104 & .990 & 6 & \citealp{hendrycks2021measuring} \\
91 & math500 & .061 & .592 & [.390, .794] & .276 & .818 & 6 & \citealp{lightman2023let} \\
92 & lsat\_qa & .062 & .584 & [.216, .952] & .026 & 1.000 & 6 & \citealp{zhong2021ar} \\
93 & natural\_qa\_openbook\_longans & .067 & .558 & [.317, .799] & .293 & .857 & 6 & \citealp{kwiatkowski2019natural} \\
94 & flores\_en\_th & .073 & .568 & [-.149, 1.285] & .346 & .894 & 3 & \citealp{costajussa2022no} \\
95 & flores\_en\_id & .074 & .569 & [-.013, 1.151] & .309 & .764 & 3 & \citealp{costajussa2022no} \\
96 & winogrande & .077 & .506 & [.248, .765] & .146 & .919 & 6 & \citealp{sakaguchi2019winogrande} \\
97 & copyright\_text & .081 & .611 & [.295, .927] & .236 & .948 & 6 & \citealp{helm2023} \\
98 & financial\_scenarios & .081 & .575 & [-.663, 1.812] & .000 & .886 & 3 & \citealp{helm2023}$^\dagger$ \\
99 & facts\_grounding & .082 & .612 & [.315, .909] & .207 & .922 & 6 & \citealp{cheng2025facts} \\
100 & musr & .086 & .476 & [.305, .646] & .273 & .644 & 6 & \citealp{sprague2023musr} \\
101 & indonli & .089 & .584 & [-.156, 1.323] & .301 & .894 & 3 & \citealp{mahendra2021indonli} \\
102 & raft & .093 & .616 & [.289, .943] & .065 & .948 & 6 & \citealp{alex2021raft} \\
103 & flores\_en\_ta & .095 & .590 & [-.192, 1.371] & .379 & .951 & 3 & \citealp{costajussa2022no} \\
104 & mlhsd & .097 & .592 & [.143, 1.040] & .434 & .789 & 3 & \citealp{susanto2025sea}$^\dagger$ \\
105 & bigcodebench & .100 & .596 & [-.288, 1.480] & .285 & .984 & 3 & \citealp{zhuo2024bigcodebench} \\
106 & boolq & .108 & .630 & [.329, .932] & .130 & .974 & 6 & \citealp{clark2019boolq} \\
107 & entity\_data\_imputation & .126 & .649 & [.377, .920] & .244 & .893 & 6 & \citealp{helm2023} \\
108 & narrative\_qa & .130 & .621 & [.392, .851] & .366 & .922 & 6 & \citealp{kocisky2017narrativeqa} \\
109 & dyck\_language & .138 & .660 & [.387, .933] & .398 & 1.000 & 6 & \citealp{helm2023} \\
110 & banking77 & .138 & .632 & [-.134, 1.399] & .278 & .846 & 3 & \citealp{casanueva2020efficient} \\
111 & financebench & .143 & .636 & [-.615, 1.887] & .055 & .943 & 3 & \citealp{islam2023financebench} \\
112 & bold & .149 & .679 & [.398, .960] & .146 & .870 & 6 & \citealp{dhamala2021bold} \\
113 & math & .155 & .545 & [.180, .909] & .130 & .992 & 6 & \citealp{sobhani2025mathmist} \\
114 & entity\_matching & .155 & .678 & [.400, .956] & .236 & .987 & 6 & \citealp{helm2023} \\
115 & the\_pile & .159 & .689 & [.328, 1.049] & .081 & 1.000 & 6 & \citealp{gao2020pile} \\
116 & hellaswag & .162 & .599 & [.329, .870] & .252 & .961 & 6 & \citealp{zellers2019hellaswag} \\
117 & fin\_qa & .164 & .658 & [-.324, 1.641] & .210 & .959 & 3 & \citealp{chen2021finqa} \\
118 & imdb & .168 & .690 & [.382, .998] & .288 & .948 & 6 & \citealp{maas2011learning} \\
119 & thaitoxicitytweets & .171 & .666 & [.098, 1.234] & .502 & .927 & 3 & \citealp{susanto2025sea}$^\dagger$ \\
120 & flores\_id\_en & .176 & .671 & [-.289, 1.631] & .228 & .935 & 3 & \citealp{costajussa2022no} \\
121 & flores\_vi\_en & .228 & .723 & [-.273, 1.720] & .260 & .967 & 3 & \citealp{costajussa2022no} \\
122 & civil\_comments & .241 & .764 & [.564, .964] & .528 & 1.000 & 6 & \citealp{borkan2019nuanced} \\
123 & flores\_ta\_en & .268 & .763 & [.514, 1.013] & .667 & .867 & 3 & \citealp{costajussa2022no} \\
124 & summarization\_cnndm & .325 & .855 & [.710, 1.001] & .618 & .987 & 6 & \citealp{hermann2015teaching} \\
\end{longtable}
\endgroup

\clearpage

\section{Top benchmarks per factor}\label{top-benchmarks-per-factor}

For each of the six solutions in
\hyperref[tab:point-summaries]{Table~\ref*{tab:point-summaries}}, the table
lists the five benchmarks with the largest absolute loading on the general
factor (\(G\)) and on each group factor of the Schmid--Leiman solution. The full
loadings are released with the code.

\begin{longtable}{@{}>{\raggedright\arraybackslash}p{0.16\textwidth}l>{\raggedright\arraybackslash}p{0.72\textwidth}@{}}
\caption{Five benchmarks with the largest absolute loading on each factor, with loadings in parentheses.}\label{tab:factor-top-benchmarks}\\
\toprule
Solution & Factor & Top benchmarks (loading) \\
\midrule
\endfirsthead
\caption[]{(continued)}\\
\toprule
Solution & Factor & Top benchmarks (loading) \\
\midrule
\endhead
\bottomrule
\endlastfoot
missForest, S Std. & $G$ & synthetic\_\allowbreak{}reasoning (+0.99), babi\_\allowbreak{}qa (+0.99), pwc\_\allowbreak{}arc\_\allowbreak{}easy (+0.97), quac (+0.97), pwc\_\allowbreak{}piqa (+0.96) \\
 & F1 & --- (all $|\lambda| < 0.01$) \\
 & F2 & indonli (+0.95), xnli (+0.92), thai\_\allowbreak{}exam\_\allowbreak{}a\_\allowbreak{}level (+0.90), xcopa (+0.90), nusax (+0.88) \\
 & F3 & kaggle\_\allowbreak{}andrewmingwang\_\allowbreak{}facts (+0.77), kaggle\_\allowbreak{}sripalthilakraj\_\allowbreak{}global\_\allowbreak{}mmlu\_\allowbreak{}lite\_\allowbreak{}english (+0.74), kaggle\_\allowbreak{}aminmohamedmohami\_\allowbreak{}browsecomp (+0.72), kaggle\_\allowbreak{}yulongt\_\allowbreak{}facts\_\allowbreak{}parametric (+0.70), kaggle\_\allowbreak{}andrewmingwang\_\allowbreak{}scicode\_\allowbreak{}main\_\allowbreak{}standard (+0.70) \\
 & F4 & twitter\_\allowbreak{}aae (+0.87), disinformation\_\allowbreak{}reiteration (-0.86), real\_\allowbreak{}toxicity\_\allowbreak{}prompts (-0.84), disinformation\_\allowbreak{}wedging (-0.82), blimp (+0.75) \\
 & F5 & legal\_\allowbreak{}support (-0.73), lsat\_\allowbreak{}qa (-0.72), gsm8k (+0.42), civil\_\allowbreak{}comments (-0.41), thai\_\allowbreak{}exam\_\allowbreak{}tgat (+0.41) \\
\midrule
k-NN, C Std. & $G$ & mmlu\_\allowbreak{}pro (+0.82), arena\_\allowbreak{}hard\_\allowbreak{}auto (+0.81), chatbot\_\allowbreak{}arena (+0.81), mmlu (+0.79), bbq (+0.79) \\
 & F1 & thai\_\allowbreak{}exam\_\allowbreak{}tpat1 (+0.63), thaiexam (+0.60), thai\_\allowbreak{}exam\_\allowbreak{}tgat (+0.59), thai\_\allowbreak{}exam\_\allowbreak{}a\_\allowbreak{}level (+0.58), facts\_\allowbreak{}grounding (-0.39) \\
 & F2 & entity\_\allowbreak{}data\_\allowbreak{}imputation (+0.79), narrative\_\allowbreak{}qa (+0.70), raft (+0.69), imdb (+0.67), natural\_\allowbreak{}qa\_\allowbreak{}openbook\_\allowbreak{}longans (+0.61) \\
 & F3 & arc (+0.73), hellaswag (+0.72), winogrande (+0.59), mmlu (+0.54), gsm8k (+0.48) \\
 & F4 & aratrust (+0.88), arabicmmlu (+0.86), madinah\_\allowbreak{}qa (+0.84), arabic\_\allowbreak{}exams (+0.82), alghafa (+0.80) \\
 & F5 & the\_\allowbreak{}pile (-0.63), gpqa\_\allowbreak{}diamond (+0.49), kaggle\_\allowbreak{}aminmohamedmohami\_\allowbreak{}browsecomp (+0.47), artificial\_\allowbreak{}analysis\_\allowbreak{}intelligence (+0.45), summarization\_\allowbreak{}cnndm (-0.39) \\
 & F6 & math\_\allowbreak{}chain\_\allowbreak{}of\_\allowbreak{}thought (+0.66), gsm (+0.65), lsat\_\allowbreak{}qa (+0.61), synthetic\_\allowbreak{}reasoning\_\allowbreak{}natural (+0.47), quac (+0.43) \\
 & F7 & civil\_\allowbreak{}comments (+0.49), pwc\_\allowbreak{}piqa (-0.40), real\_\allowbreak{}toxicity\_\allowbreak{}prompts (-0.32), pwc\_\allowbreak{}arc\_\allowbreak{}challenge (-0.32), babi\_\allowbreak{}qa (-0.30) \\
 & F8 & aime25 (+0.78), kaggle\_\allowbreak{}andrewmingwang\_\allowbreak{}scicode\_\allowbreak{}subproblem\_\allowbreak{}with\_\allowbreak{}background (+0.60), kaggle\_\allowbreak{}andrewmingwang\_\allowbreak{}simpleqa\_\allowbreak{}verified (+0.60), kaggle\_\allowbreak{}andrewmingwang\_\allowbreak{}scicode\_\allowbreak{}subproblem\_\allowbreak{}standard (+0.59), kaggle\_\allowbreak{}sjmikler\_\allowbreak{}livecodebench (+0.54) \\
\midrule
missForest, C Std. & $G$ & thai\_\allowbreak{}exam\_\allowbreak{}tpat1 (+0.93), bbq (+0.93), anthropic\_\allowbreak{}red\_\allowbreak{}team (+0.92), simple\_\allowbreak{}safety\_\allowbreak{}tests (+0.92), thai\_\allowbreak{}exam\_\allowbreak{}tgat (+0.92) \\
 & F1 & bbq (+0.33), legalbench (+0.31), xstest (+0.30), pwc\_\allowbreak{}arc\_\allowbreak{}challenge (+0.30), arena\_\allowbreak{}hard\_\allowbreak{}auto (+0.30) \\
 & F2 & summarization\_\allowbreak{}cnndm (+0.96), imdb (+0.92), entity\_\allowbreak{}data\_\allowbreak{}imputation (+0.86), summarization\_\allowbreak{}xsum (+0.83), narrative\_\allowbreak{}qa (+0.82) \\
 & F3 & aime25 (+0.96), kaggle\_\allowbreak{}sjmikler\_\allowbreak{}livecodebench (+0.90), artificial\_\allowbreak{}analysis\_\allowbreak{}intelligence (+0.89), madinah\_\allowbreak{}qa (+0.83), arabic\_\allowbreak{}exams (+0.83) \\
 & F4 & real\_\allowbreak{}toxicity\_\allowbreak{}prompts (-0.61), disinformation\_\allowbreak{}wedging (-0.60), disinformation\_\allowbreak{}reiteration (-0.58), kaggle\_\allowbreak{}vijitsingh1\_\allowbreak{}mgsm\_\allowbreak{}english (+0.40), bold (-0.36) \\
\midrule
SoftImpute, C Std. & $G$ & gsm (+0.73), math\_\allowbreak{}chain\_\allowbreak{}of\_\allowbreak{}thought (+0.72), openbookqa (+0.72), mbpp (+0.66), bbh (+0.65) \\
 & F1 & arabicmmlu (-0.95), madinah\_\allowbreak{}qa (-0.95), aratrust (-0.95), arabic\_\allowbreak{}exams (-0.93), alghafa (-0.92) \\
 & F2 & math (+0.81), ifeval (+0.70), chatbot\_\allowbreak{}arena (+0.67), mbpp (+0.62), synthetic\_\allowbreak{}reasoning\_\allowbreak{}natural (+0.60) \\
 & F3 & math500 (+0.88), disinformation\_\allowbreak{}reiteration (-0.83), real\_\allowbreak{}toxicity\_\allowbreak{}prompts (-0.83), disinformation\_\allowbreak{}wedging (-0.83), kaggle\_\allowbreak{}andrewmingwang\_\allowbreak{}scicode\_\allowbreak{}subproblem\_\allowbreak{}standard (+0.77) \\
 & F4 & narrative\_\allowbreak{}qa (+1.00), imdb (+0.95), natural\_\allowbreak{}qa\_\allowbreak{}openbook\_\allowbreak{}longans (+0.81), entity\_\allowbreak{}data\_\allowbreak{}imputation (+0.80), raft (+0.77) \\
 & F5 & hellaswag (+0.98), arc (+0.94), winogrande (+0.83), mmlu (+0.81), artificial\_\allowbreak{}analysis\_\allowbreak{}intelligence (-0.56) \\
 & F6 & simple\_\allowbreak{}safety\_\allowbreak{}tests (+0.95), anthropic\_\allowbreak{}red\_\allowbreak{}team (+0.94), xstest (+0.94), bbq (+0.73), harmbench (+0.66) \\
 & F7 & legal\_\allowbreak{}support (+0.90), lsat\_\allowbreak{}qa (+0.86), copyright\_\allowbreak{}text (-0.67), wmt\_\allowbreak{}14 (+0.54), summarization\_\allowbreak{}xsum (+0.43) \\
 & F8 & kaggle\_\allowbreak{}aminmohamedmohami\_\allowbreak{}browsecomp (-0.79), bfcl (+0.55), multiloko (-0.51), arena\_\allowbreak{}hard\_\allowbreak{}auto (+0.46), legalbench (+0.45) \\
 & F9 & bold (+0.77), facts\_\allowbreak{}grounding (-0.65), copyright\_\allowbreak{}text (+0.51), civil\_\allowbreak{}comments (-0.47), entity\_\allowbreak{}matching (+0.47) \\
 & F10 & gsm (+0.61), openbookqa (+0.61), math\_\allowbreak{}chain\_\allowbreak{}of\_\allowbreak{}thought (+0.53), gsm8k (+0.42), wmt\_\allowbreak{}14 (+0.33) \\
\midrule
SoftImpute, R Aggr. & $G$ & financial\_\allowbreak{}scenarios (+0.90), madinah\_\allowbreak{}qa (+0.90), kaggle\_\allowbreak{}yulongt\_\allowbreak{}facts\_\allowbreak{}parametric (+0.90), sorry\_\allowbreak{}bench (-0.90), pwc\_\allowbreak{}siqa (+0.89) \\
 & F1 & pwc\_\allowbreak{}peerqa (+0.70), include (+0.44), mt\_\allowbreak{}bench (+0.41), naturalquestions (+0.39), agentif (+0.38) \\
 & F2 & litbench (+0.39), chw\_\allowbreak{}care\_\allowbreak{}plan (-0.34), vihsd (-0.33), head\_\allowbreak{}qa (-0.33), benchmax (+0.32) \\
 & F3 & cogbench (+0.45), scigen (+0.44), worldvaluesbench (+0.42), ewok\_\allowbreak{}social\_\allowbreak{}relations (+0.42), ewok\_\allowbreak{}agent\_\allowbreak{}properties (+0.42) \\
 & F4 & summarization\_\allowbreak{}cnndm (+0.79), summarization\_\allowbreak{}xsum (+0.62), narrative\_\allowbreak{}qa (+0.50), pwc\_\allowbreak{}arc\_\allowbreak{}challenge (+0.47), pwc\_\allowbreak{}arc\_\allowbreak{}easy (+0.45) \\
 & F5 & kaggle\_\allowbreak{}jonlipovetz\_\allowbreak{}game\_\allowbreak{}arena (-0.75), oogiri (-0.72), kaggle\_\allowbreak{}jonlipovetz\_\allowbreak{}chess\_\allowbreak{}suite (-0.67), SuperGPQA (-0.64), workplacehumor (+0.50) \\
 & F6 & medi\_\allowbreak{}qa (-0.86), facts\_\allowbreak{}parametric (+0.80), medication\_\allowbreak{}qa (-0.62), math\_\allowbreak{}regular (+0.57), medmcqa (-0.49) \\
 & F7 & bold (+0.89), copyright\_\allowbreak{}text (+0.85), pwc\_\allowbreak{}bioasq (+0.58), pwc\_\allowbreak{}lambada (+0.54), twitter\_\allowbreak{}aae (-0.47) \\
 & F8 & pwc\_\allowbreak{}bioasq (-0.91), indoculture (+0.81), mega (+0.54), ice (-0.43), indommlu (+0.42) \\
 & F9 & pubmedqa (-1.09), naturalquestions (+0.70), pwc\_\allowbreak{}svamp (-0.53), pwc\_\allowbreak{}big\_\allowbreak{}bench\_\allowbreak{}penguins\_\allowbreak{}in\_\allowbreak{}a\_\allowbreak{}table (+0.52), pwc\_\allowbreak{}big\_\allowbreak{}bench\_\allowbreak{}causal\_\allowbreak{}judgment (+0.52) \\
 & F10 & ehr\_\allowbreak{}sql (+0.86), medbullets (+0.71), medmcqa (+0.53), medec (+0.49), pwc\_\allowbreak{}frontiermath (+0.49) \\
 & F11 & flores\_\allowbreak{}vi\_\allowbreak{}en (+0.82), flores\_\allowbreak{}id\_\allowbreak{}en (+0.69), filbench (+0.53), wisesight (+0.53), flores\_\allowbreak{}en\_\allowbreak{}id (+0.51) \\
 & F12 & disinformation\_\allowbreak{}wedging (-0.53), disinformation\_\allowbreak{}reiteration (-0.50), twitter\_\allowbreak{}aae (+0.40), msmarco\_\allowbreak{}trec (-0.38), msmarco\_\allowbreak{}regular (-0.38) \\
 & F13 & pwc\_\allowbreak{}obqa (-0.72), pwc\_\allowbreak{}record (+0.46), pwc\_\allowbreak{}storycloze (-0.43), irokobench (+0.33), triviaqa (+0.31) \\
 & F14 & legal\_\allowbreak{}support (+0.45), lsat\_\allowbreak{}qa (+0.34), aratrust (+0.24), livebench (+0.23), hellaswag (-0.22) \\
 & F15 & vmlu (+1.14), indommlu (+0.60), medqa (+0.55), cogbench (+0.48), legalbench (+0.40) \\
 & F16 & boolq (-0.68), babi\_\allowbreak{}qa (-0.62), civil\_\allowbreak{}comments (-0.61), entity\_\allowbreak{}data\_\allowbreak{}imputation (-0.60), entity\_\allowbreak{}matching (-0.40) \\
 & F17 & mbpp (-0.35), mtrag (-0.35), sea\_\allowbreak{}exam (-0.35), gpqa (+0.34), pwc\_\allowbreak{}pecc (-0.34) \\
 & F18 & complexbench (-0.35), pwc\_\allowbreak{}arc\_\allowbreak{}challenge (-0.33), disinformation\_\allowbreak{}reiteration (-0.32), neuro\_\allowbreak{}eval (+0.31), harmbench (-0.31) \\
 & F19 & artificial\_\allowbreak{}analysis\_\allowbreak{}intelligence (-0.38), livecodebench (-0.29), aime\_\allowbreak{}2025 (-0.27), swe\_\allowbreak{}bench (-0.23), gpqa\_\allowbreak{}diamond (-0.22) \\
 & F20 & indicqa (+0.55), arabic\_\allowbreak{}exams (+0.38), pwc\_\allowbreak{}sst\_\allowbreak{}5\_\allowbreak{}fine\_\allowbreak{}grained\_\allowbreak{}classification (+0.36), tydiqa (+0.34), arabicmmlu (+0.33) \\
\midrule
SoftImpute, S Std. & $G$ & twitter\_\allowbreak{}aae (-0.72), alghafa (-0.69), aratrust (-0.66), ice (-0.65), arabic\_\allowbreak{}exams (-0.64) \\
 & F1 & blimp (-0.69), ice (-0.65), twitter\_\allowbreak{}aae (-0.62), alghafa (-0.49), the\_\allowbreak{}pile (-0.48) \\
 & F2 & msmarco\_\allowbreak{}regular (+1.00), msmarco\_\allowbreak{}trec (+0.98), boolq (+0.84), wikifact (+0.77), entity\_\allowbreak{}data\_\allowbreak{}imputation (+0.74) \\
 & F3 & gsm (+0.95), math\_\allowbreak{}chain\_\allowbreak{}of\_\allowbreak{}thought (+0.94), openbookqa (+0.91), medqa (+0.84), math\_\allowbreak{}regular (+0.78) \\
 & F4 & kaggle\_\allowbreak{}andrewmingwang\_\allowbreak{}scicode\_\allowbreak{}subproblem\_\allowbreak{}with\_\allowbreak{}background (+0.86), mmlu\_\allowbreak{}prox (+0.86), kaggle\_\allowbreak{}andrewmingwang\_\allowbreak{}scicode\_\allowbreak{}main\_\allowbreak{}with\_\allowbreak{}background (+0.85), math500 (+0.81), kaggle\_\allowbreak{}andrewmingwang\_\allowbreak{}scicode\_\allowbreak{}subproblem\_\allowbreak{}standard (+0.80) \\
 & F5 & mlhsd (+0.97), vihsd (+0.95), thaitoxicitytweets (+0.91), uitvsfc (+0.87), lindsea\_\allowbreak{}pragmatics\_\allowbreak{}presuppositions\_\allowbreak{}id (+0.84) \\
 & F6 & simple\_\allowbreak{}safety\_\allowbreak{}tests (+0.95), anthropic\_\allowbreak{}red\_\allowbreak{}team (+0.93), xstest (+0.72), summarization\_\allowbreak{}cnndm (+0.71), arena\_\allowbreak{}hard\_\allowbreak{}auto (+0.65) \\
 & F7 & xquad (-0.75), bbh (+0.69), kaggle\_\allowbreak{}aminmohamedmohami\_\allowbreak{}browsecomp (-0.67), lsat\_\allowbreak{}qa (+0.66), musr (+0.64) \\
 & F8 & disinformation\_\allowbreak{}reiteration (+0.85), disinformation\_\allowbreak{}wedging (+0.83), artificial\_\allowbreak{}analysis\_\allowbreak{}intelligence (+0.74), real\_\allowbreak{}toxicity\_\allowbreak{}prompts (+0.67), gpqa\_\allowbreak{}diamond (+0.56) \\
\end{longtable}

\clearpage

\section{Common-Subject UMAP Plots}\label{common-subject-distances}

This section presents the UMAP plots for the benchmarks' composite distance,
with each figure showing the plots of one imputation method. Each plot is shown
for one collapse strategy. The aggregate plot is composited over all collapse
strategies and all imputation methods. This section has 8 figures in total.

Notably, through the different imputation strategies, all of the visualizations
show the same pattern, in which shared-domain benchmarks tend to be dispersed
across the vector space. Highlighting the same benchmarks as in the results
(with some benchmarks discarded by each densifier), we see the same pattern
that, for example, \texttt{aime25} and \texttt{gsm8k} are distanced quite far
from each other. Through most of the different possible solutions, our core
claim that same-domain benchmarks are not guaranteed to cluster together is
robust.

\begin{figure}
\centering
\pandocbounded{\includegraphics[keepaspectratio]{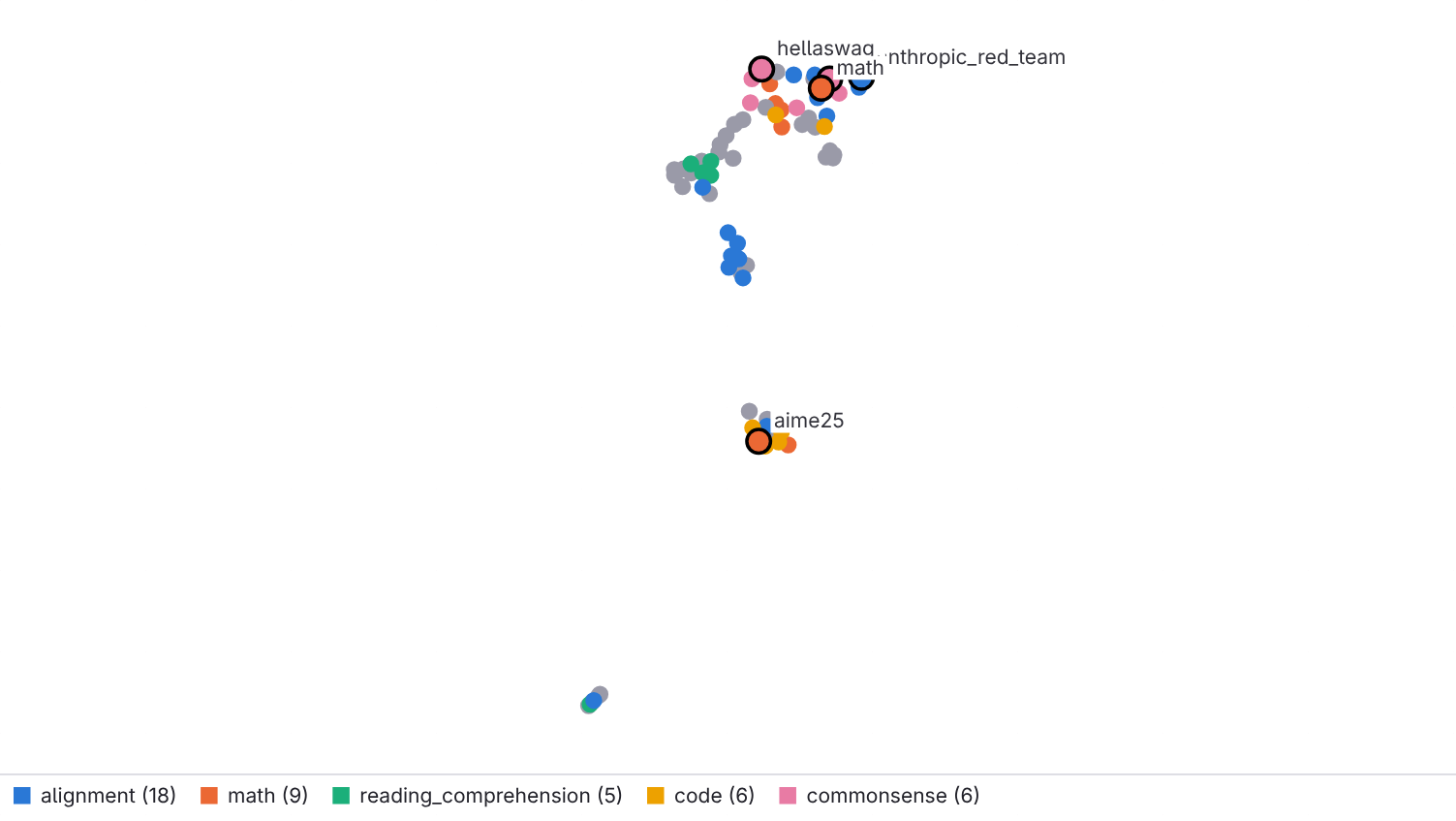}}
\caption{C dataset, aggregated across imputations.}
\end{figure}

\begin{figure}
\centering
\pandocbounded{\includegraphics[keepaspectratio]{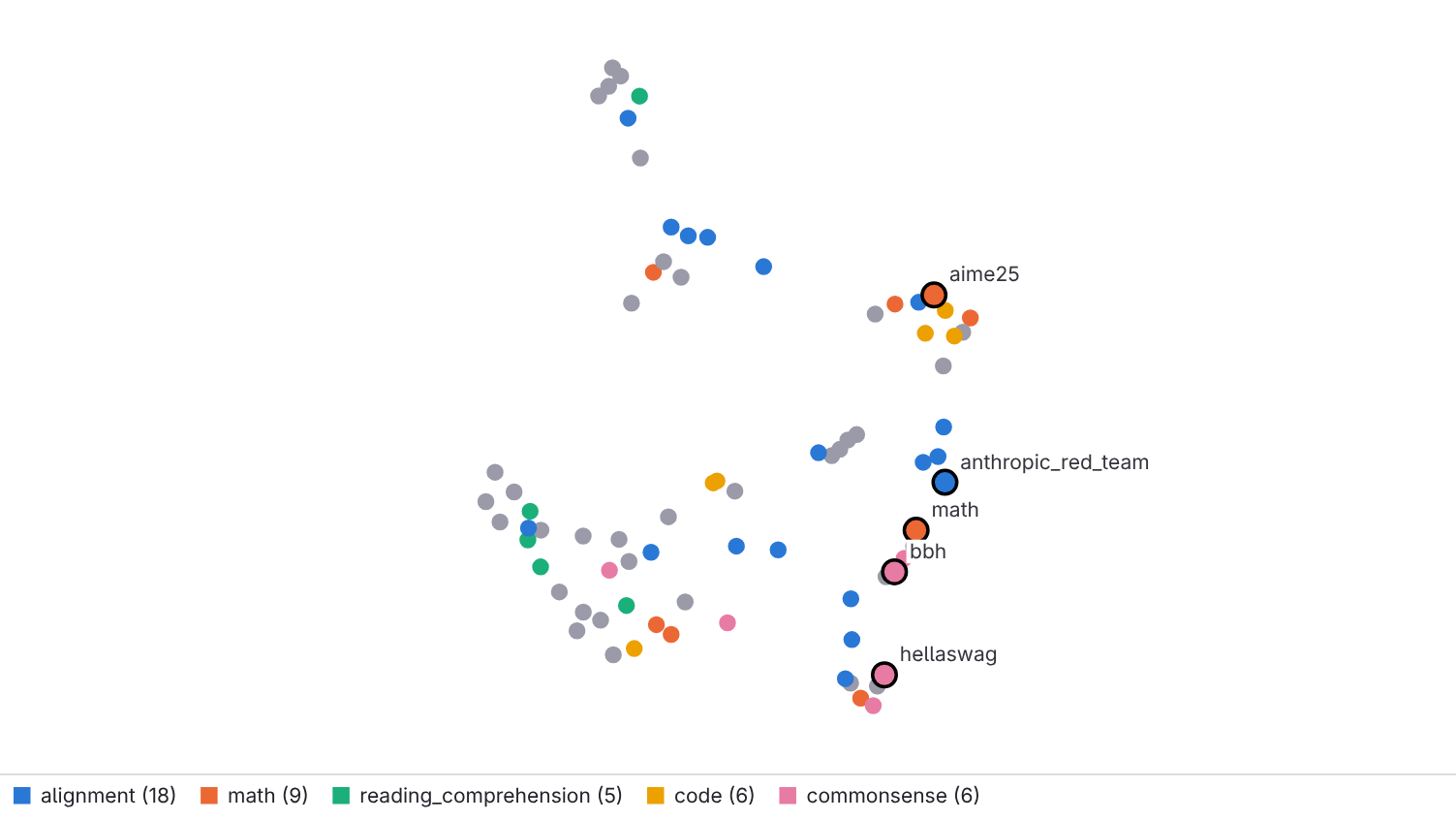}}
\caption{C Std., k-NN (\(wR^2 = 0.442\)).}
\end{figure}

\begin{figure}
\centering
\pandocbounded{\includegraphics[keepaspectratio]{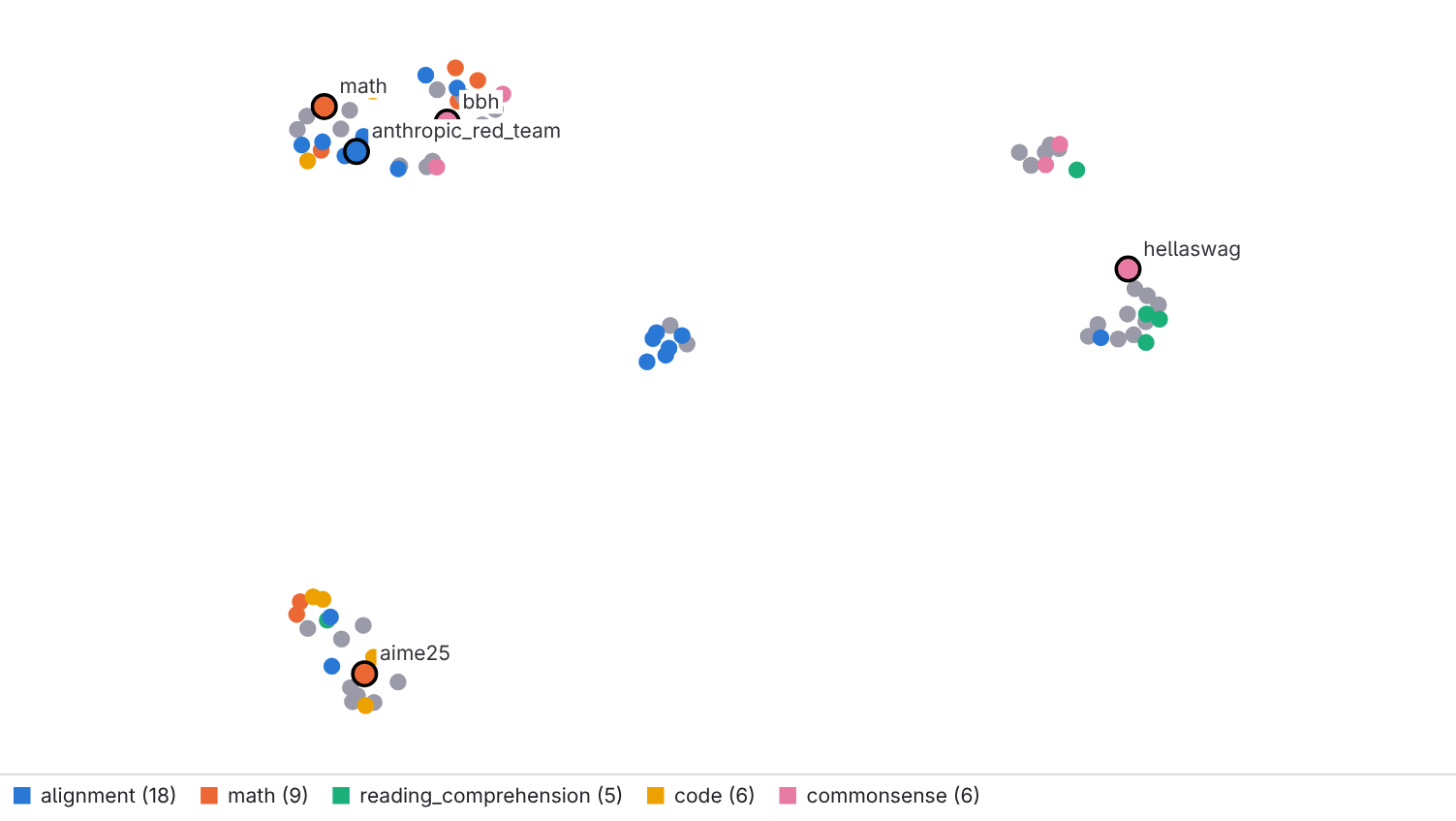}}
\caption{C Std., missForest (\(wR^2 = 0.566\)).}
\end{figure}

\begin{figure}
\centering
\pandocbounded{\includegraphics[keepaspectratio]{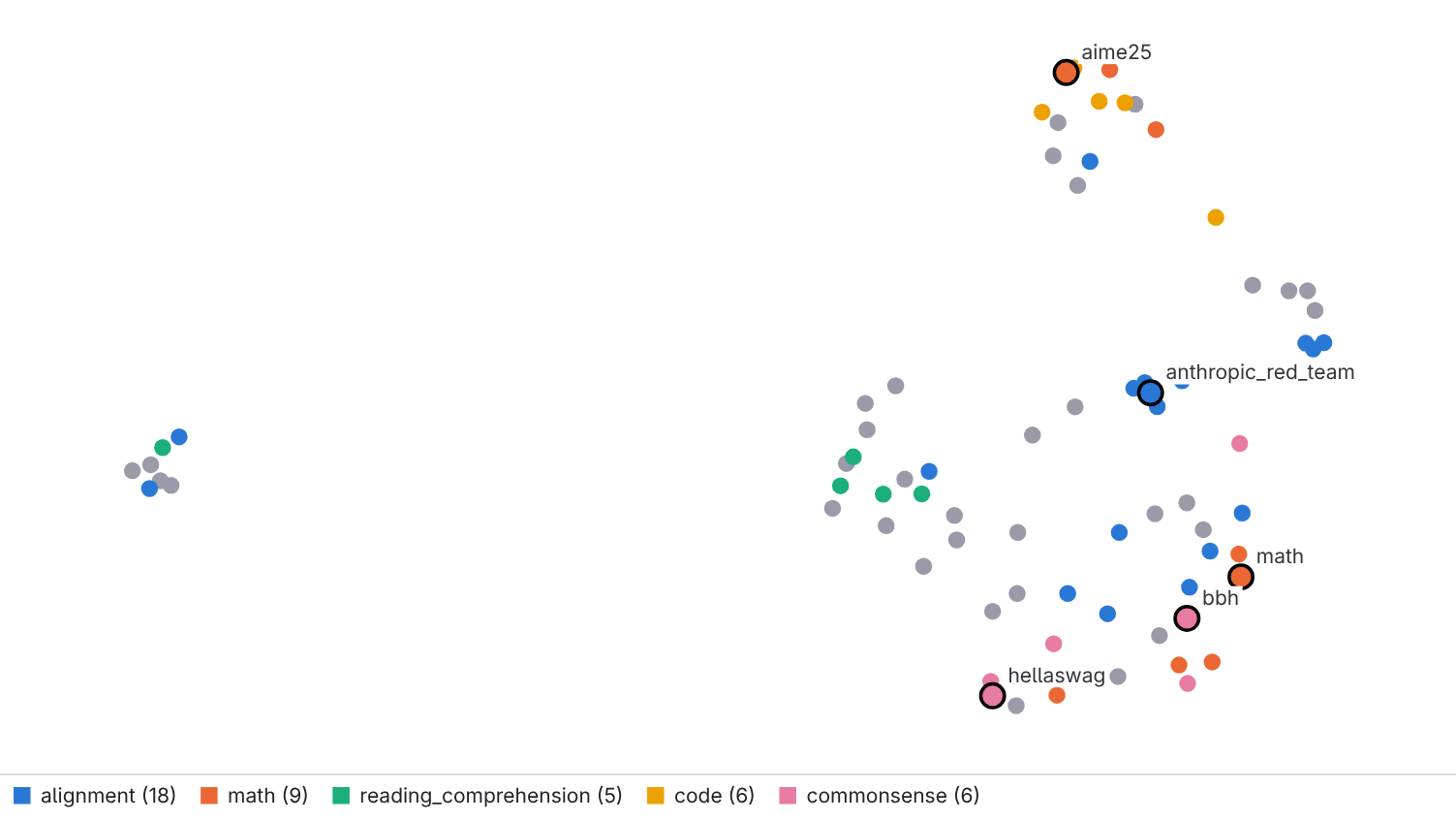}}
\caption{C Std., SoftImpute (\(wR^2 = 0.586\)).}
\end{figure}

\begin{figure}
\centering
\pandocbounded{\includegraphics[keepaspectratio]{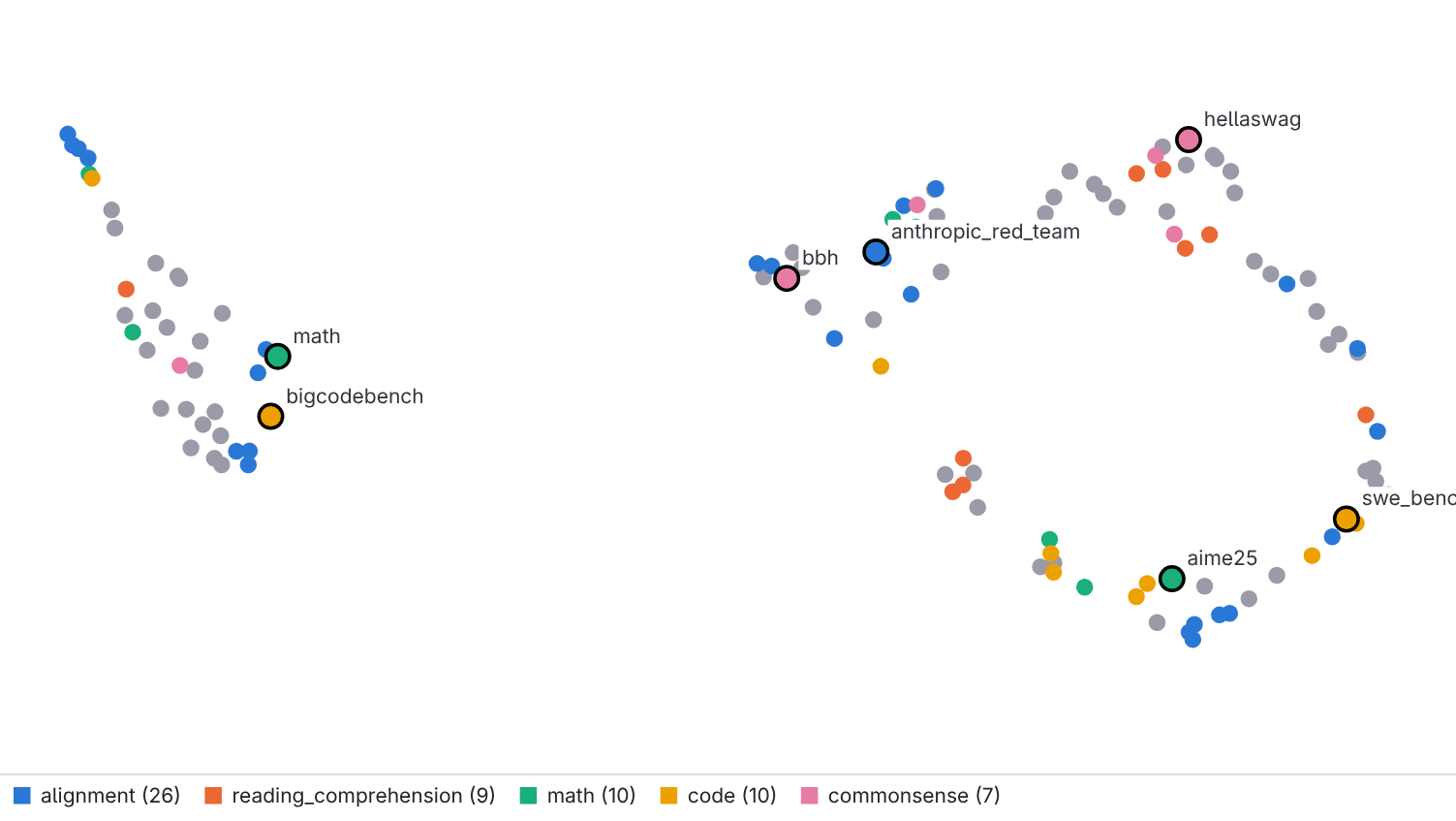}}
\caption{S Std., missForest (\(wR^2 = 0.455\)).}
\end{figure}

\begin{figure}
\centering
\pandocbounded{\includegraphics[keepaspectratio]{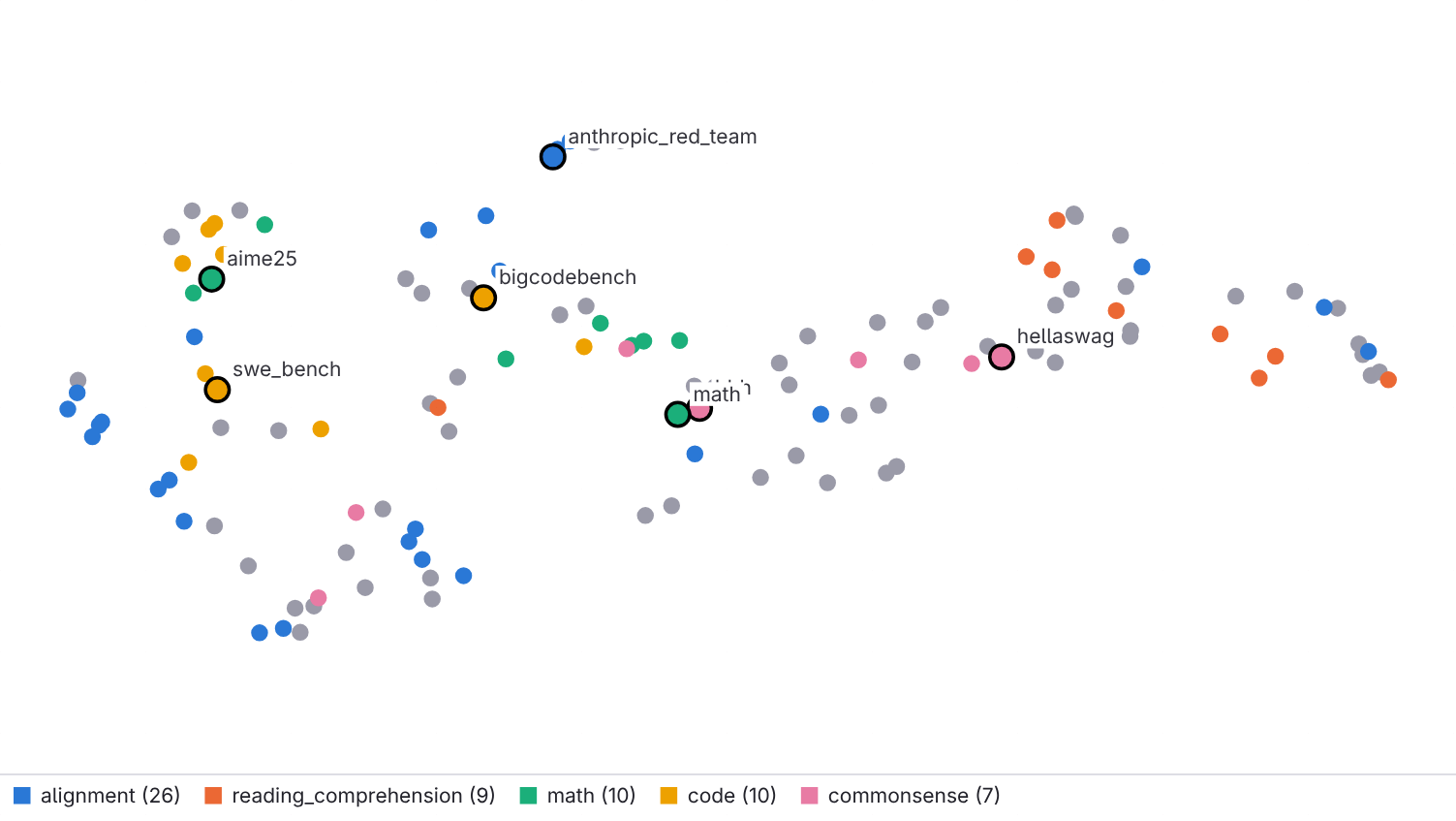}}
\caption{S Std., SoftImpute (\(wR^2 = 0.428\)).}
\end{figure}

\begin{figure}
\centering
\pandocbounded{\includegraphics[keepaspectratio]{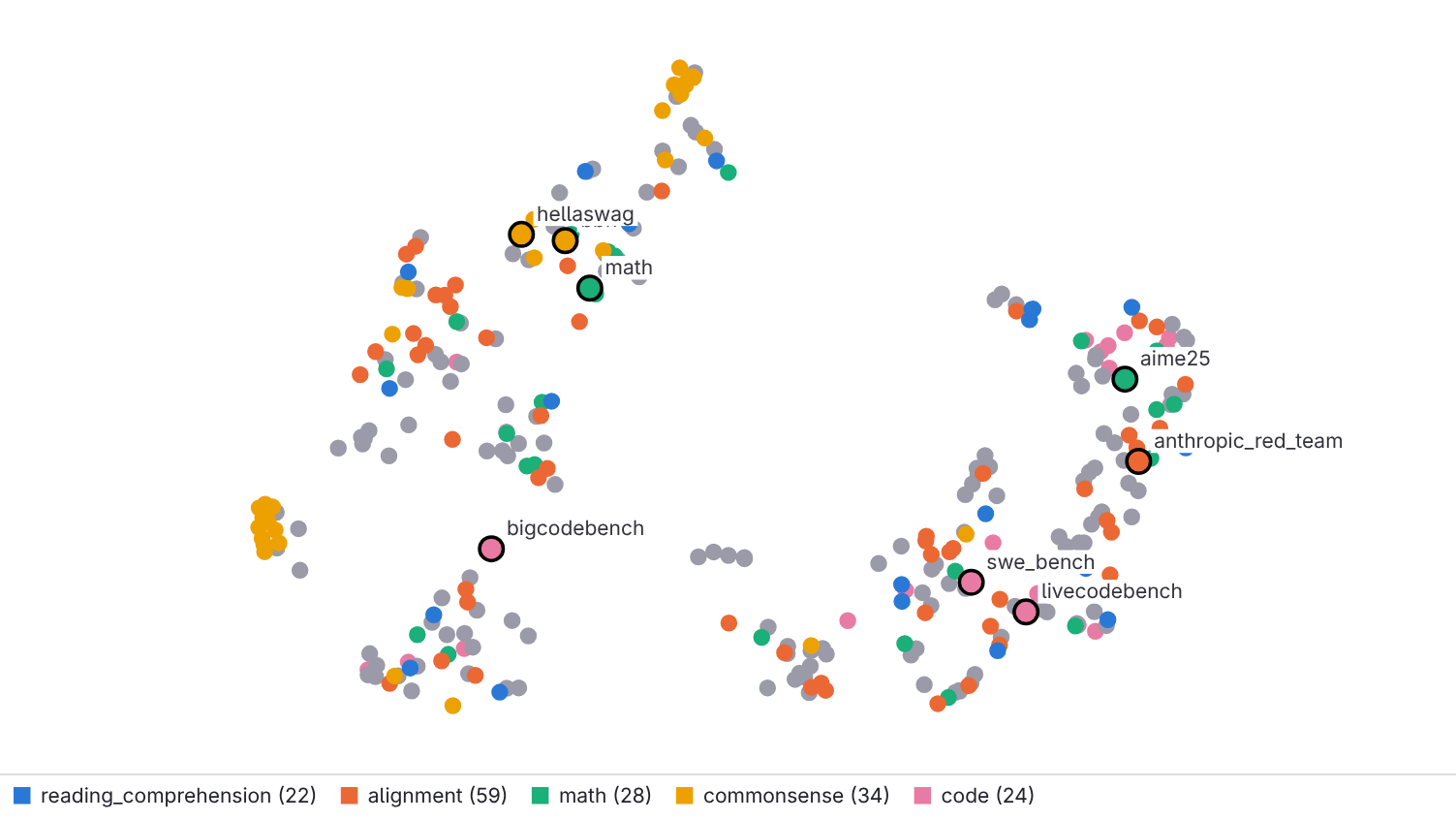}}
\caption{R Aggr.}
\end{figure}

\begin{figure}
\centering
\pandocbounded{\includegraphics[keepaspectratio]{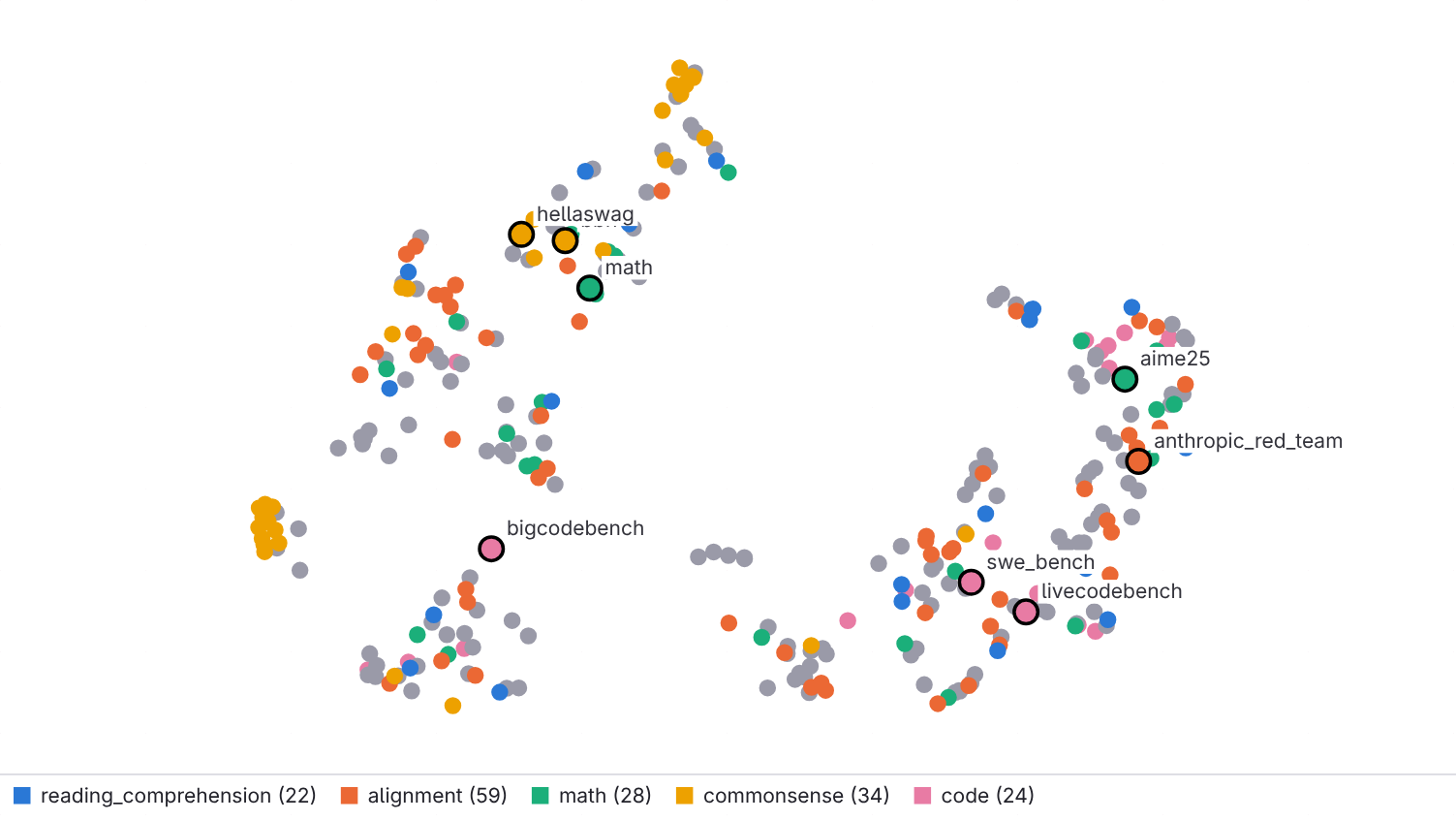}}
\caption{R Aggr., SoftImpute (\(wR^2 = 0.441\)).}
\end{figure}


\end{document}